\documentclass[10pt,twocolumn,letterpaper]{article}

\usepackage[pagenumbers]{iccv} 

\definecolor{iccvblue}{rgb}{0.21,0.49,0.74}
\usepackage[pagebackref,breaklinks,colorlinks,allcolors=iccvblue]{hyperref}

\def\paperID{11758} 
\def\confName{ICCV}
\def\confYear{2025}

\usepackage[utf8]{inputenc} 
\usepackage[T1]{fontenc}    
\usepackage{hyperref}       
\usepackage{url}            
\usepackage{booktabs}       
\usepackage{amsfonts}       
\usepackage{nicefrac}       
\usepackage{microtype}      
\usepackage{xcolor}         
\definecolor{umn_maroon}{RGB}{122, 0, 25}
\definecolor{myred}{HTML}{D62728}
\definecolor{myblue}{HTML}{1F77B4}
\definecolor{mygreen}{HTML}{00FF00}
\definecolor{darkpastelgreen}{rgb}{0.01, 0.75, 0.24}

\usepackage{graphicx,amsmath,amsfonts,amsthm,amscd,amssymb,bm,url,color,latexsym}
\usepackage{physics}
\allowdisplaybreaks
\usepackage[capitalize,nameinlink]{cleveref}

\usepackage{bbold}
\renewcommand{\mathbf}{\boldsymbol}

\newcommand{\mb}{\mathbf}
\newcommand{\mc}{\mathcal}

\newcommand{\bb}{\mathbb}

\newcommand{\paren}{\pqty}

\newcommand{\ol}{\overline}

\newcommand{\T}{\intercal}

\usepackage{wrapfig}
\usepackage{enumitem}
\usepackage{algorithm}%
\usepackage{algpseudocode}

\usepackage{siunitx}
\usepackage{multirow}
\usepackage{tikz}
\usepackage{sidecap}
\usepackage{arydshln}
\usetikzlibrary{decorations.pathreplacing,calc}
\newcommand{\tikzmark}[1]{\tikz[overlay,remember picture] \node (#1) {};}
\newcommand*{\AddNote}[4]{%
    \begin{tikzpicture}[overlay, remember picture]
        \draw [decoration={brace,amplitude=0.5em},decorate,ultra thick,blue]
            ($(#3)!(#1.north)!($(#3)-(0,1)$)$) --  
            ($(#3)!(#2.south)!($(#3)-(0,1)$)$)
                node [align=left, text width=2.5cm, pos=0.5, anchor=west] {#4};
    \end{tikzpicture}
}%
\usepackage{amssymb}%
\usepackage{pifont}%
\usepackage{empheq}

\title{Temporal-Consistent Video Restoration with Pre-trained Diffusion Models}

\author{
Hengkang Wang$^1$\thanks{This work of Wang H. was partially done while interning at Amazon.com, Inc.} \quad Yang Liu$^2$ \quad Huidong Liu$^2$ \quad Chien-Chih Wang$^2$ \quad Yanhui Guo$^2$ \\
Hongdong Li$^{2,3}$ \quad Bryan Wang$^2$  \quad Ju Sun$^1$\\
$^1$Computer Science and Engineering, University of Minnesota {\tt\small \{wang9881,jusun\}@umn.edu} \\
$^2$Amazon.com, Inc. {\tt\small \{yliuu,liuhuido,ccwang,yanhuig,hongdli,brywan\}@amazon.com} \quad $^3$ANU
}

\begin{document}
\maketitle
\begin{abstract}
Video restoration (VR) aims to recover high-quality videos from degraded ones. Although recent zero-shot VR methods using pre-trained diffusion models (DMs) show good promise, they suffer from approximation errors during reverse diffusion and insufficient temporal consistency. Moreover, dealing with 3D video data, VR is inherently computationally intensive. In this paper, we advocate viewing the reverse process in DMs as a function and present a novel Maximum a Posterior (MAP) framework that directly parameterizes video frames in the seed space of DMs, eliminating approximation errors. We also introduce strategies to promote bilevel temporal consistency: semantic consistency by leveraging clustering structures in the seed space, and pixel-level consistency by progressive warping with optical flow refinements. Extensive experiments on multiple virtual reality tasks demonstrate superior visual quality and temporal consistency achieved by our method compared to the state-of-the-art.

\end{abstract}   
    
\section{Introduction}
\label{sec:intro}

\begin{figure*}[t]
    \centering
    \includegraphics[width=0.85\textwidth]{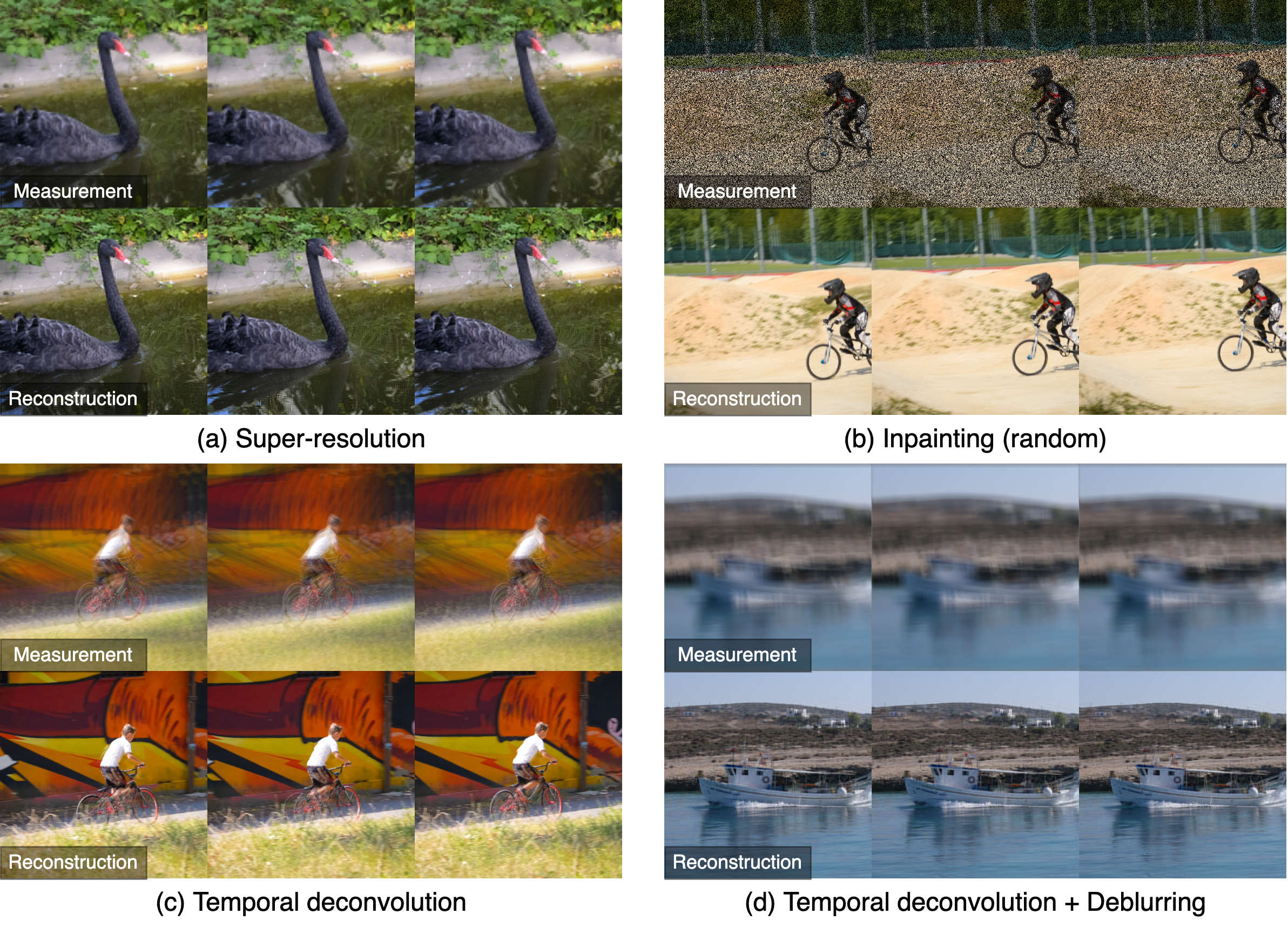}
    \caption{Visualization of sample VR results by our method: (a) Super-resolution $\times 4$; (b) Inpainting with $50\%$ random pixel masking; (c) Temporal deconvolution using uniform PSF with kernel width $k = 7$; and (d) Temporal deconvolution with motion deblurring.}
    \label{fig:our_res}
    \vspace{-1em}
\end{figure*}

Video restoration (VR) aims to recover high-quality (HQ) videos \(\mb X\) from given low-quality (LQ) observations \(\mb Y \approx \mc A \paren{\mb X}\), where \(\mc A\) represents a spatial and/or temporal degradation. Typical VR tasks include super-resolution~\cite{zhou_upscale--video_2023,chan_investigating_2021,wang_exploiting_2023,liang_vrt_2022}, inpainting~\cite{zhou_propainter_2023,lugmayr_repaint_2022,xu_deep_2019}, and deblurring~\cite{zhong_efficient_2020,nah_ntire_2019}. 


Modern VR methods rely on deep learning and fall into two main categories. \textbf{(1) The supervised learning approach} trains deep neural networks (DNNs) on LQ-HQ paired data, i.e., \(\{(\mathbf{Y}_i, \mathbf{X}_i)\}_{i=1}^{K}\), to learn direct mappings from \(\mathbf{Y}\) to \(\mathbf{X}\). Although conceptually simple, these methods require large-scale, high-quality paired datasets and significant computational resources (e.g. \emph{32 A100-80G GPUs} for super-resolution~\cite{zhou_upscale--video_2023}). Moreover, they need to train new DNNs for different VR tasks, with limited task adaptability; \textbf{(2)} \textbf{The zero-shot paradigm} enabled by pre-trained deep generative models, especially diffusion models (DMs)~\cite{ho_denoising_2020,song_denoising_2022}. Due to the lack of mature video DMs, recent research~\cite{yeh_diffir2vr-zero_2024, cao_zero-shot_2024, kwon_solving_2024, kwon_vision-xl_2024} has adopted pre-trained image DMs for VR, achieving remarkable results without task-specific retraining. 

Most zero-shot DM-based VR methods~\cite{kwon_solving_2024,cao_zero-shot_2024,kwon_vision-xl_2024,yeh_diffir2vr-zero_2024} interleave reverse diffusion steps with iterative gradient updates or projections to approach the feasible set \(\{\mathbf{X} \mid \mathbf{Y} \approx \mathcal{A}(\mathbf{X})\}\). However, these methods usually face three fundamental challenges. First, they suffer from \textbf{unavoidable approximation errors} \textbf{\textcolor{blue}{(Challenge 1)}} when approximating an intractable likelihood~\cite{chung_diffusion_2023}, regardless of whether they employ gradient updates or projections. These errors accumulate throughout the reverse diffusion process, potentially degrading the reconstruction quality. Second, these zero-shot methods often struggle for satisfactory \textbf{temporal consistency} \textbf{\textcolor{blue}{(Challenge 2)}} due to the difficulty in extracting accurate motion information from LQ measurements and the absence of explicit motion priors in the image DMs they leverage. Third, compared to image restoration, VR entails \textbf{significantly more computation and memory footprints} \textbf{\textcolor{blue}{(Challenge 3)}}, creating substantial efficiency barriers that must be addressed for practical applications. 

\textbf{In this paper, we focus on solving video restoration (VR) problems using pre-trained image DMs}, while addressing the three fundamental challenges faced by state-of-the-art (SOTA) methods. We specifically choose \textbf{latent diffusion models (LDMs)} as our backbone DMs due to their superior generation quality, computational efficiency, and widespread adoption. \textbf{\textcolor{blue}{(Tackling Challenge 1)}} Our method builds upon the classical Maximum a Posterior (MAP) framework~\cite{ulyanov_deep_2017,pan_exploiting_2020,zhuang_blind_2022,li_deep_2023,li_random_2023,wang_dmplug_2024}:
\begin{align}  \label{eq:MAP}
    \min_{\mb X}\ \; \underbrace{\ell\paren{\mb Y, \mc{A}\paren{\mb X}}}_{\text{data consistency}} + \lambda \underbrace{\Omega\paren{\mb X}}_{\text{regularization}}.  
\end{align}
We take a novel perspective that views the entire reverse diffusion process as a function $\mc R$ which, when composed with the pre-trained decoder $\mc D$ in LDMs, maps from the seed space directly to the image manifold $\mc M$. This allows us to naturally reparameterize the video frame-by-frame as $\mb X = [\mc D \circ \mc R\paren{\mb z_1}, \dots, \mc D \circ \mc R\paren{\mb z_N}]$; plugging this into \cref{eq:MAP} leads to a unified optimization formulation \textbf{with respect to the seeds $\mb Z = [\mb z_1, \dots, \mb z_N]$}. This reparametrization approach effectively leverages the powerful LDM priors for VR while avoiding the approximation errors inherent in SOTA methods. To \textbf{\textcolor{blue}{address Challenge 2}}, we design a hierarchical framework to promote bilevel temporal consistency. For \textbf{semantic-level temporal consistency}, we first explore the seed space and observe an intriguing \textit{clustering phenomenon}: input seeds of frames from different videos are scattered, while those of frames within the same video are clustered. Motivated by this, we construct a noise prior by hypothesizing that consecutive frames share a common seed with only minor frame-specific variations. To enhance \textbf{pixel-level temporal consistency}, we implement a progressive warping mechanism that combines image warping with incremental optical flow (OF) refinements. Our ablation study in \cref{tab:ablation_stages} underscores the effectiveness of both components in enhancing bilevel temporal consistency. To \textbf{\textcolor{blue}{deal with Challenge 3}}, we introduce an efficient diffusion sampling strategy using the DDIM sampler. Notably, our ablation study (\cref{tab:ablation_steps}) reveals that \(4\) reverse steps in $\mc R$ are sufficient to achieve the SOTA performance. Moreover, by leveraging the multivariate mean value theorem, we significantly reduce the computational complexity of the proposed method. To further boost efficiency, we make the trainable residuals for each frame more lightweight through low-rank approximations. \textbf{Our contributions} can be summarized as follows:
\begin{itemize}[nosep]
    \item We propose a MAP-based framework for VR that harnesses pre-trained image DMs by reparameterizing frames via the entire reverse diffusion process, eliminating the approximation errors that have plagued SOTA methods. 
    
    \item We devise a compelling hierarchical approach for bilevel temporal consistency that unites semantic-level coherence (through our key discovery of clustering patterns in the seed space) with pixel-level precision (via dynamic progressive warping with optical flow refinements).
    
    \item We design our method with exceptional computational efficiency through three innovations: an optimized DDIM sampling strategy that requires only $4$ steps, an approximate reformulation using the multivariate mean value theorem, and memory-efficient trainable residuals with low-rank approximations.
    
    \item Our comprehensive experiments on challenging VR tasks demonstrate that our method leads to substantial performance improvements over SOTA methods, providing exceptional visual quality and temporal consistency.
\end{itemize}

\section{Background and related work}
\label{sec:bg}

\paragraph{\textcolor{umn_maroon}{Diffusion models (DMs)}}  
Recently, DMs have dominated generative models, capable of producing high-quality objects. The early denoising diffusion probabilistic model (\textbf{DDPM})~\cite{ho_denoising_2020} involves two processes: a \emph{forward diffusion} process that transforms any clean data sample $\mb x_0 \sim p_{\text{data}}$  into pure noise $\mb x_T \sim \mc N\paren{\mb 0, \mb I}$ by sequential noise injection, governed by the stochastic differential equation (SDE): $d\mb x = -\beta_t/2 \cdot \mb x dt + \sqrt{\beta_t} d\mb w$, where $\beta_t$ represents the noise schedule, and $\mb w$ denotes the standard Wiener process; a \emph{reverse diffusion} process that performs sequential denoising, turning any seed noise into a useful data sample---hence responsible for data generation, and follows the form: 
\begin{align} \label{eq:reverse_SDE}  
    d\mb x = -\beta_t \left[\mb x/2 + \nabla_{\mb x} \log p_t(\mb x) \right] dt + \sqrt{\beta_t} d\ol{\mb w},  
\end{align}  
where $\ol{\mb w}$ is the time-reversed Brownian motion, and the term $\nabla_{\mb x} \log p_t(\mb x)$, known as the score function, represents the gradient of the log-likelihood. To train an DM,  this score function is approximated using a DNN, {\small $\mb \varepsilon_{\mb \theta}^{(t)}\paren{\mb x}$}, trained via score matching techniques~\cite{hyvarinen_estimation_2005,song_generative_2020}.  
In practice, the diffusion process is discretized in $T$ time steps, using a predefined variance schedule $\beta_1, \dots, \beta_T$. Defining $\alpha_t \doteq 1 - \beta_t$ with $\alpha_T \to 0$ and $\bar{\alpha}_t \doteq \prod_{s=1}^t \alpha_s$, the forward steps are written as: $\mb x_t = \sqrt{1-\beta_t} \mb x_{t-1} + \sqrt{\beta_t} \mb z$, where $\mb z \sim \mc N\paren{\mb 0, \mb I}$. The reverse steps run as $\mb x_{t-1} = 1/\sqrt{\alpha_t} \cdot  \left(\mb x_t - \beta_t \mb \varepsilon_{\mb \theta}^{(t)}(\mb x_t) / \sqrt{1-\bar{\alpha}_t} \right) + \sqrt{\beta_t} \mb z$; 
this iterative sampling process usually requires a large number of steps to achieve high-quality generation, leading to slow inference. To address this slowness, the denoising diffusion implicit model (\textbf{DDIM})~\cite{song_denoising_2022} introduces a non-Markovian relaxation of the forward process, allowing each step $\mb x_t$ to depend not only on $\mb x_{t-1}$ but also directly on $\mb x_0$.
This relaxation enables sampling with fewer steps while maintaining generation quality. The reverse step in DDIM is  
\begin{align}\label{eq:ddim}  
    \mb x_{t-1} = \sqrt{\bar{\alpha}_{t-1}} \widehat{\mb x}_0\paren{\mb x_t} + \sqrt{1-\bar{\alpha}_{t-1}} \mb \varepsilon_{\mb \theta}^{(t)}(\mb x_t),  
\end{align}  
where $\widehat{\mb x}_0\paren{\mb x_t} \doteq (\mb x_t - \sqrt{1-\bar{\alpha}_t} \mb \varepsilon_{\mb \theta}^{(t)}(\mb x_t))/\sqrt{\bar{\alpha}_t}$ estimates the clean image $\mb x_0$ from $\mb x_t$. 

Despite DDIM's speedup, training diffusion models in high-resolution pixel spaces remains computationally expensive. Latent diffusion models (\textbf{LDMs})~\cite{rombach_high-resolution_2021} overcome this by performing both training and inference in low-dimensional latent spaces, wrapped into pre-trained encoder-decoder models. The LDM framework has become dominant in SOTA visual generative models~\cite{rombach_high-resolution_2021,podell_sdxl_2023,peebles_scalable_2023}. 

\vspace{-1em}
\paragraph{\textcolor{umn_maroon}{DMs for video restoration}} 

Approaches to VR using DMs can be categorized into two main classes: supervised~\cite{daras_warped_2024,zhou_upscale--video_2023} and zero-shot~\cite{kwon_solving_2024,kwon_vision-xl_2024,cao_zero-shot_2024,yeh_diffir2vr-zero_2024}. Supervised methods train DM-based models on paired data or correlated noise, which is outside our focus (see discussion \cref{sec:intro}). Zero-shot methods largely inherit ideas from zero-shot image restoration with pre-trained DMs and most of them focus on directly modeling the conditional distribution $p_t(\mb x| \mb y)$ and substitute the unconditional score function $\nabla_{\mb x} \log p_t(\mb x)$ in \cref{eq:reverse_SDE} with the conditional score function $\nabla_{\mb x} \log p_t(\mb x | \mb y) =  \nabla_{\mb x} \log p_t(\mb x) + \nabla_{\mb x} \log p_t(\mb y | \mb x)$, i.e.,
\begin{multline} \label{eq:cond_reverse_sde}
d\mb x = -\beta_t\left[\mb x/2 + \paren{\nabla_{\mb x} \log p_t(\mb x) + \nabla_{\mb x} \log p_t(\mb y | \mb x)} \right] dt \\
+ \sqrt{\beta_t} d \ol{\mb w}.
\end{multline}
Here, while $\nabla_{\mb x} \log p_t(\mb x)$ can be approximated using the pre-trained score function $\mb \varepsilon_{\mb \theta}^{(t)}\paren{\mb x}$, the term $\nabla_{\mb x} \log p_t(\mb y | \mb x)$ remains intractable because $\mb y$ does not depend directly on $\mb x(t)$. To circumvent this, one line of research directly approximates $p_t(\mb y | \mb x(t))$~\cite{chung_diffusion_2023,fei_generative_2023}, while the other interleaves diffusion reverse steps from \cref{eq:ddim} with projections (or gradient updates)~\cite{chung_decomposed_2023} to guide the process toward the feasible set $\{\mb x | \mb y \approx \mc A \paren{\mb x}\}$. Unfortunately, both strategies introduce \textbf{unavoidable approximation errors} \textbf{\textcolor{blue}{(Challenge 1)}} that can accumulate during the reverse diffusion process (RDP), ultimately limiting their practical performance. For VR, \cite{cao_zero-shot_2024} follows the generative diffusion prior (GDP) framework~\cite{fei_generative_2023} using gradient steps, while \cite{kwon_solving_2024,kwon_vision-xl_2024} implement the decomposed diffusion sampling (DDS)~\cite{chung_decomposed_2023} approach with several conjugate gradient (CG) update steps. However, \textit{CG requires the degradation operator $\mathcal{A}$ to be known, symmetric and positive-definite}~\cite{shewchuk_introduction_1994}, restricting its applicability for VR. 

VR also needs good temporal consistencies. To address this issue, existing zero-shot methods use batch-consistent sampling strategies~\cite{kwon_solving_2024,kwon_vision-xl_2024} and leverage optical flow (OF) guidance. However, accurate OF estimation from LQ videos is inherently challenging. Different approaches attempt to overcome this limitation: \cite{yeh_diffir2vr-zero_2024} directly obtains OF from LQ videos and proposes a hierarchical latent warping technique, while \cite{cao_zero-shot_2024} acquires OFs during intermediate stages of the RDP---when results may still contain noise---and applies these OFs for warping in the image space. In summary, the SOTA methods achieve only semantic-level alignment~\cite{kwon_solving_2024,kwon_vision-xl_2024,yeh_diffir2vr-zero_2024} or attempt pixel-level alignment with suboptimal OFs~\cite{cao_zero-shot_2024}, leaving \textbf{temporal consistency} in VR a critical standing challenge \textbf{\textcolor{blue}{(Challenge 2)}}.

\section{Our method}
\label{sec:method}

In this paper, we employ \textbf{pre-trained image LDMs} as priors to solve video restoration (VR) problems due to their superior generation quality, computational efficiency, and widespread adoption. In \cref{subsec:formulation}, we propose a novel LDM-based formulation that overcomes the limitations of SOTA methods, addressing \textbf{\textcolor{blue}{challenge 1}}. In \cref{subsec:temporal}, we introduce two effective components for bilevel temporal consistency, tackling \textbf{\textcolor{blue}{challenge 2}}. Finally, in \cref{subsec:ingredients}, we present several essential techniques to ensure computational and memory efficiency, targeting \textbf{\textcolor{blue}{challenge 3}}.

\subsection{Our basic formulation}
\label{subsec:formulation}

To mitigate approximation errors in existing interleaving methods, we introduce a novel and principled formulation for solving VR problems following the classical maximum a posteriori (MAP) principle. Our goal is to recover a high-quality video $\mb X$ that not only satisfies the measurement constraint $\mb Y \approx \mc A(\mb X)$, but also resides near the realistic video manifold $\mc M$: $\min_{\mb X \in \mc M}\ \; \ell\paren{\mb Y, \mc{A}\paren{\mb X}}$. Inspired by \cite{wang_dmplug_2024}, we propose viewing the entire RDP as a function $\mc R$, which, when composed with the pre-trained decoder $\mc D$ in LDMs, maps the seed space to the video space. This fresh perspective enables us to reparameterize the video as $\mb X = \mc D \circ \mc R(\mb Z)$, which can then be plugged into the MAP framework \cref{eq:MAP}, leading to a unified formulation:
\begin{align} \label{eq:ours1}
    \mb Z^* \in \min\nolimits_{\mb Z}\; \ell\paren{\mb Y, \mc{A} \paren{\mc D \circ \mc R\paren{\mb Z}}}, 
\end{align}
where $\mb Z = [\mb z_1, \dots, \mb z_N]$, and $\mc D$ and $\mc R$ are applied framewise. The final reconstruction can be obtained as $\mb X^* = \mc D \circ \mc{R}(\mb Z^*)$. Note that the RDP consists of multiple iterative steps. Mathematically, a single reverse step can be expressed as a function $g$ that depends on {\small $\mb \varepsilon_{\mb \theta}^{(i)}$}, i.e., {\small $g_{\mb \varepsilon_{\mb \theta}^{(i)}}$}, representing the $i$-th reverse step that maps the latent variable $\mb z_{i+1}$ to $\mb z_{i}$. The full RDP can then be written as  
\begin{align} \label{eq:R_def}
    \mc R \doteq g_{\mb \varepsilon_{\mb \theta}^{(0)}} \circ g_{\mb \varepsilon_{\mb \theta}^{(1)}} \circ \cdots \circ g_{\mb \varepsilon_{\mb \theta}^{(T-2)}} \circ g_{\mb \varepsilon_{\mb \theta}^{(T-1)}}.
\end{align}

\subsection{Promoting bilevel temporal consistency} \label{subsec:temporal}

Although our formulation in \cref{eq:ours1} can address VR tasks leveraging pre-trained image LDMs, the reconstructed videos still may not have good temporal consistency (see ``Base'' in \cref{tab:ablation_stages}). This is not surprising, as it tries to recover individual frames separately and fails to capture inter-frame dependencies---a critical issue to be addressed by all methods relying on image DMs~\cite{cao_zero-shot_2024,kwon_solving_2024,kwon_vision-xl_2024,yeh_diffir2vr-zero_2024}. 
In this section, we address this critical issue \textbf{\textcolor{blue}{(Tackling Challenge 2)}} by introducing two distinct components that target temporal consistency at two different levels.

\subsubsection{Noise prior for semantic-level consistency}
\label{subsec:semantic}

\begin{figure}[!htpb]
\centering
\includegraphics[width=0.45\textwidth]{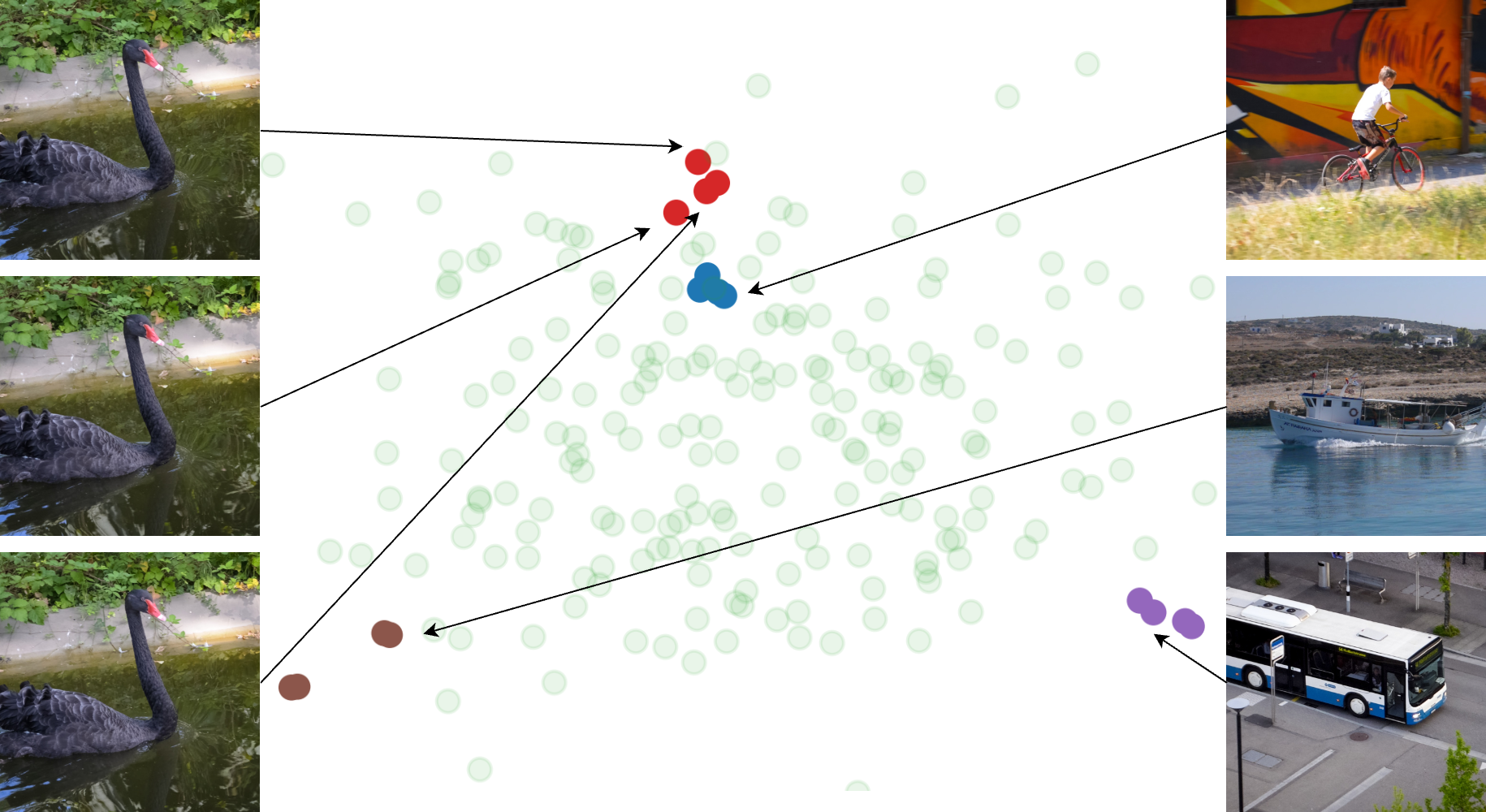}
\caption{T-SNE~\cite{maaten_visualizing_2008} visualization of seeds extracted for video frames. Green dots are i.i.d. Gaussian noise. Seeds for the same video form clusters, while those for different videos are scattered.}
\label{fig:cluster}
\vspace{-1em}
\end{figure}

\paragraph{\textcolor{umn_maroon}{An intriguing clustering phenomenon}}

To strengthen temporal consistency, we begin by exploring potential structures in the seed space. We obtain seeds $\mb Z$ for randomly sampled frames from videos by solving the regression problem: $\min\nolimits_{\mb{Z}}\; \ell\paren{\mb{X}, \mc{D} \circ \mc{R}(\mb{Z})}$. From \cref{fig:cluster}, we observe an interesting pattern: seeds for frames of different videos are widely scattered without apparent correlation, whereas \textbf{seeds for frames of the same video tend to cluster together}---a pattern we can potentially leverage to improve temporal consistency across frames. 

To this end, we hypothesize that consecutive frames share a common seed but have minor frame-specific deviations. Accordingly, we decompose the seed matrix as $\mb Z = \mb z_{shared} \bb 1^\T + \mb R$, where $\mb z_{shared}$ represents the shared seed, $\mb R = [\mb r_1, \dots, \mb r_N]$ captures frame-specific residuals, and $\bb 1$ is an all-one vector. Moreover, we need to ensure that the residuals are small by constraining $\|\mb r_i\|_2 \leq \sigma$ for a small constant $\sigma > 0$. Our new formulation built on \cref{eq:ours1} is 
\begin{align} \label{eq:ours2}
\begin{split}
  \min\nolimits_{\mb z_{shared}, \mb R} & \quad \ell\paren{\mb{Y}, \mc{A}\paren{\mc{D} \circ {\mc{R}\paren{\mb z_{shared} \bb 1^\T+ \mb R}}}}, \\
  \text{s.t.} & \quad \|\mb r_i\|_2 \leq \sigma, \quad \forall i \in \{1, \ldots, N\}, 
  \end{split}
\end{align}
where $\mc{D}$ and $\mc{R}$ are applied framewise. To solve this constrained optimization problem, we choose the Projected Gradient Descent (PGD) method. 
Our algorithm alternates between gradient descent and projection: 
\begin{align}
\begin{split}
(\mb{z}_{shared}, \mb{R}) & \leftarrow (\mb{z}_{shared}, \mb{R}) - \eta \nabla_{(\mb{z}_{shared}, \mb{R})} \ell, \\
\mb{r}_i & \leftarrow \Pi_{\|\mb{r}_i\|_2 \leq \sigma}(\mb{r}_i), \quad \forall i \in \{1, \ldots, N\}. 
\end{split}
\end{align}
Here, $\Pi_{\|\mb{r}_i\|_2 \leq \sigma}$ denotes the projection operator that maps $\mb{r}_i$ to the closest point within the $\ell_2$ norm ball of radius $\sigma$. 

\subsubsection{Progressive warping for pixel-level consistency}
\label{subsec:pixel}

As shown in \cref{tab:ablation_stages}, incorporating the noise prior improves both data fidelity and temporal consistency. However, fine-grained consistency across frames remains weak. To enhance it, we introduce a progressive warping loss:
\begin{align} \label{eq:warp} 
& \mc{L}_{\text{warp}} = \sum\nolimits_{n=1}^{N-1} \ell\left(\mathbf{M} \odot \mathbf{x}'_n, \mathbf{M} \odot \mathcal{W}\left(\mathbf{x}'_{n+1}, \mathbf{f}'_{n \rightarrow n+1}\right)\right), \nonumber \\ 
& \mathbf{f}'_{n \rightarrow n+1} = \text{stopgrad}\left(\text{RAFT} \left(\mathbf{x}'_n, \mathbf{x}'_{n+1}\right)\right),
\end{align}
where $\mathcal{W}$ denotes backward warping~\cite{sun_pwc-net_2017}, $\mathbf{M}$ is the estimated non-occlusion mask obtained via forward-backward consistency checks~\cite{meister_unflow_2017}, and $\mathbf{f}'_{n \rightarrow n+1}$ represents the optical flow (OF) from frame $n$ to $n+1$ and is treated as constant during backpropagation. The frame reconstructed at the time step $n$ is denoted as $\mathbf{x}'_n$. We compute the OF using the pre-trained estimator RAFT~\cite{teed_raft_2020}. This warping loss explicitly penalizes pixel-level changes between motion-compensated consecutive frames, thereby enhancing temporal consistency at the pixel level.

\begin{figure}[!htpb]
\vspace{-1em}
\centering
\includegraphics[width=0.5\textwidth]{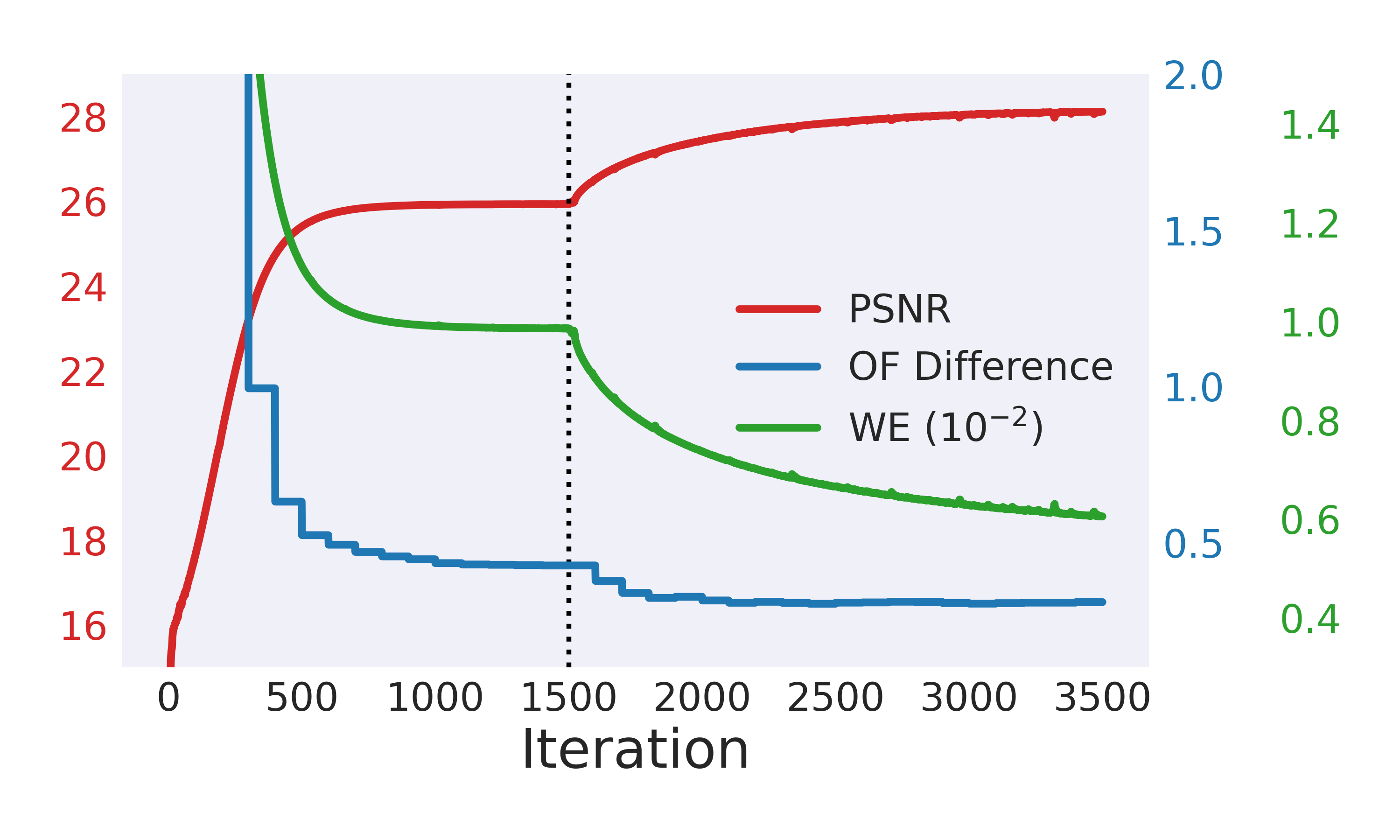}
\vspace{-2.5em}
\caption{Evolution of key metrics during the iterative VR process: PSNR, OF difference (measured against ground truth), and warping error (WE). All metrics improve in early iterations. After OF difference stabilizes (vertical dotted line), progressive warping is activated, further enhancing PSNR and reducing WE, demonstrating our framework's effectiveness for temporal consistency.}
\label{fig:evolution}
\vspace{-1em}
\end{figure}
In practice, we introduce the warping loss $\mathcal{L}_{\text{warp}}$ only after the estimated frames $\mathbf{x}'_1, \dots, \mathbf{x}'_N$ attain sufficient quality, as early iterations often yield noisy results. The evolution of performance over iterations is illustrated in \cref{fig:evolution}; it shows an evident performance boost in both data fidelity and temporal consistency after incorporating the proposed warping loss. To reduce computation, we update the OFs every $P$ iterations instead of each. We term our approach \textbf{progressive warping} because the estimated OFs themselves are progressively refined during iterations. However, these estimated OFs can exhibit slight fluctuations, which can hinder stable training. To address this, we apply an exponential moving average (EMA) to smooth out the OFs:  
\begin{equation}
\mathbf{f}^{(t)}_{n \rightarrow n+1} = \beta \mathbf{f}^{(t-1)}_{n \rightarrow n+1} + (1-\beta) \mathbf{f}'^{(t)}_{n \rightarrow n+1},
\end{equation}
where $\mathbf{f}^{(t)}_{n \rightarrow n+1}$ is the stabilized flow at iteration $t$, $\mathbf{f}'^{(t)}_{n \rightarrow n+1}$ is the newly estimated flow, and $\beta \in [0,1)$ is the weighted averaging coefficient. This smoothing mechanism not only stabilizes the training process, but also enhances the temporal consistency of the OFs. 

\subsection{Boosting computational and memory efficiency} \label{subsec:ingredients}

\paragraph{\textcolor{umn_maroon}{Efficient diffusion sampling}} 
DDPMs typically require dozens or hundreds of sampling steps, inducing significant computational and memory burdens for our approach in \cref{eq:ours2}. To address this challenge, we implement the DDIM sampler for the reverse process $\mc{R}$, which enables us to skip intermediate steps while preserving generation quality. Surprisingly, we find that $4$ reverse steps are sufficient for our method to outperform the SOTA, as shown in \cref{tab:ablation_steps}. Adding more steps does not provide substantial benefits and can even slightly degrade the results, possibly due to numerical issues from vanishing gradients. Hence, we take $4$ as the default number of reverse steps. 
We also implement gradient checkpointing techniques to further reduce memory costs.

\begin{table}[!htpb]
    \centering
    \caption{Ablation study on performance vs. the number of reverse steps in diffusion process $\mc R$, performed on the DAVIS dataset for video super-resolution $\times 4$.}
    \label{tab:ablation_steps}
    \resizebox{0.8\columnwidth}{!}{ 
    \begin{tabular}{c c c c c}
        \toprule
        Steps & PSNR$\uparrow$ & SSIM$\uparrow$ & LPIPS$\downarrow$ & WE($10^{-2}$)$\downarrow$ \\
        \midrule
        SOTA \cite{kwon_vision-xl_2024} & 26.03 & 0.717 & 0.339 & 1.411 \\
        \hdashline
        2 & {27.87} & {0.785} & {0.324} & {0.742} \\
        4 & \textbf{27.95} & \textbf{0.790} & \textbf{0.321} & \textbf{0.725} \\
        10 & 27.70 & 0.777 & 0.347 & 0.746 \\
        \bottomrule
    \end{tabular}
    }
\end{table}

\paragraph{\textcolor{umn_maroon}{Efficient reformulation via mean value theorem (MVT)}}

There is a potential computational bottleneck in \cref{eq:ours2}: When performing backpropagation with respect to $\mb R$, the term $\mc{R}\paren{\mb z_{shared} \bb 1^\T + \mb R}$ requires $N$ separate forward passes through $\mc R$, expensive both in computation and in memory when $N$ is large. To address this, we assume that $\mc{R}: \mathbb{R}^d \rightarrow \mathbb{R}^d$ is continuously differentiable, with the operator norm (i.e., the largest singular value) of the Jacobian uniformly bounded by $L$. Then, for all $i$, $\norm{\mc{R}\paren{\mb z_{shared} + \mb r_i} - {\mc{R}\paren{\mb z_{shared}}}} \le L \norm{\mb r_i}_2 \le L \sigma$ by the multivariate mean-value theorem~\cite{coleman2012calculus}. So we consider 
\begin{multline}\label{eq:ours3}
  \min_{\mb z_{shared}, \mb R} \; \ell\paren{\mb{Y}, \mc{A}\paren{\mc{D} \paren{ {\mc{R}\paren{\mb z_{shared}} + \mb R}}}}, \\
  \text{s.t.} \; \|\mb r_i\|_2 \leq L\sigma, \quad \forall i \in \{1, \ldots, N\}. 
\end{multline}
Since $\mb R$ is now outside the RDP $\mc R$, backpropagation through $\mc R$ is only needed for $\mb z_{shared}$, reducing both memory and computation by a factor of $N$. To further reduce the cost due to backpropagation, we repeat the above idea by decomposing the pre-trained decoder $\mc D$ as $\mc D = \mc D_1 \circ \mc D_2$ and putting the learnable residuals as input to $\mc D_1$, leading to our final formulation
\begin{empheq}[box=\fbox]{multline} \label{eq:ours4}
  \min_{\mb z_{shared}, \mb R} \; \ell\paren{\mb{Y}, \mc{A}\paren{\mc{D}_1 \paren{\mc{D}_2 \circ \mc{R}\paren{\mb z_{shared}}  + \mb R}}}, \\
  \text{s.t.} \; \|\mb r_i\|_2 \leq L'\sigma, \quad \forall i \in \{1, \ldots, N\}. 
\end{empheq}
Here, $L'$ depends on both $L$ and also the maximum operator norm of the Jacobian of $\mc D_2$, and accounts for the maximum amplification of perturbations through $\mc D_2 \circ \mc R$, ensuring rigorous bounds and applicability of our formulation to long video sequences. 
In practice, we place the trainable $\mb R$ directly on the last layer of $\mc{D}$. We set $L'\sigma$ as $C\sqrt{\mathrm{dimension}(\mb r_i)}$ for all $i$, where $C$ is a tunable hyperparameter. 

\paragraph{\textcolor{umn_maroon}{Lightweight low-rank residual parameterization}}

For VR, the trainable residuals $\mathbf{r}_1, \ldots, \mathbf{r}_N$ are high-dimensional tensors. In our implementation with Stable Diffusion, for example, the last layer of $\mathcal{D}$ has dimension $(1, 128, 512, 512)$ for each residual. This leads to substantial memory and computation burdens. We address this by enforcing low-rank structures on the residuals. Specifically, we decompose each residual along its spatial dimensions as: $\mathbf{r}_n = \mathbf{A}_n \mathbf{B}_n$, where $\mathbf{A}_n \in \mathbb{R}^{1 \times 128 \times 512 \times k}$ and $\mathbf{B}_n \in \mathbb{R}^{1 \times 128 \times k \times 512}$ with $k \ll 512$. This reduces the total parameter count for $\mb R$ from $O(N \cdot 128 \cdot 512 \cdot 512)$ to $O(N \cdot 128 \cdot 512 \cdot 2k)$, resulting in significant computational savings. To make the low-rank parameterization compatible with the PGD algorithm to solve \cref{eq:ours4}, we need to ensure $\norm{\mb A_n \mb B_n} \le L' \sigma$ for all $i$. For this, we take the heuristic projection
\begin{align}
    (\mathbf{A}_n', \mathbf{B}_n')  \leftarrow (\mathbf{A}_n, \mathbf{B}_n) / \sqrt{\|\mathbf{A}_n \mathbf{B}_n\|/L'\sigma}, 
\end{align}
after each gradient step on $(\mathbf{A}_n, \mathbf{B}_n)$. This produces a feasible $\mb r_n$, as $\norm{\mathbf{A}_n' \mathbf{B}_n'} = \norm{\mathbf{A}_n \mathbf{B}_n}/(\norm{\mathbf{A}_n \mathbf{B}_n}/L'\sigma) = L' \sigma$. With extra analytical and computational efforts, it is possible to develop a rigorous orthogonal projector here. But, we stick to this simple one, as it is easy to implement and effective in practice.  

\begin{algorithm}[!htbp]
\caption{Our video restoration framework}
\begin{algorithmic}[1]
\Require Epochs $E$, transition $E_T$, diffusion steps $T$, $\mb Y$, $\mc A$
\State Initialize $\mb z_{shared}^0 \sim \mc N\paren{\mb 0, \mb I}$ and residuals $\mb r_1^0, \cdots \mb r_N^0$
\For{$e = 0$ to $E - 1$}
    \For{$i = T - 1$ to $0$} \tikzmark{top}
    \State $\hat{\mb s} \gets \mb \varepsilon_{\mb \theta}^{(i)}\paren{\mb z_i^e}$
    \State $\hat{\mb z}_0^e \gets \frac{1}{\sqrt{\Bar{\alpha}_i}}\paren{\mb z_i^e - \sqrt{1 - \Bar{\alpha}_i} \hat{\mb s}}$
    \State $\mb z_{i-1}^e \gets$ DDIM reverse with $\hat{\mb z}_0^e, \hat{\mb s}$ \tikzmark{right}
    \EndFor \tikzmark{bottom}
    \State \#\# Current reconstruction for $n-$th frame
    \State $\mb x'_{n} = \mc D_1 \paren{\mc D_2 \circ \mc R \paren{\mb z_{shared}^e} + \mb r_{n}^e}$
    
    \If{$e >= E_T$}
        \State $\mb{f}'^{(e)}_{n \rightarrow n+1} \gets \text{stopgrad}\paren{\textbf{RAFT}(\mb{x}'_n, \mb{x}'_{n+1})}$
        \State $\mb{f}^{(e)}_{n \rightarrow n+1} = \beta \mb{f}^{(e-1)}_{n \rightarrow n+1} + (1-\beta) \mb{f}'^{(e)}_{n \rightarrow n+1}$
        \State Calculate $\mathcal{L}_{\text{warp}}$ via \cref{eq:warp}
    \Else
        \State $\mathcal{L}_{\text{warp}} = 0$
    \EndIf

    \State Update $\mb z_{shared}^{e+1}, \mb r_{1}^{e+1}, \cdots, \mb r_{N}^{e+1}$ via \cref{eq:ours4} with $\mathcal{L}_{\text{warp}}$
    \State Project each $\mb{r}_n^{e+1}$ onto $\ell_2$ norm ball
\EndFor
\Ensure Recovered video $\left[\mb {x}'_1, \dots, \mb {x}'_N \right]$, where $\mb {x}'_n = \mc D_1 \paren{\mc D_2 \circ \mc R \paren{\mb z_{shared}^E} + \mb r_{n}^E}$
\AddNote{top}{bottom}{right}{$\; \mc R$}
\end{algorithmic}
\end{algorithm}

\section{Experiments}
\label{sec:exps}


\paragraph{\textcolor{umn_maroon}{Experimental setup}} 

We conduct comprehensive experiments on five VR tasks that involve various spatial and temporal degradations, following the protocols in \cite{kwon_solving_2024,kwon_vision-xl_2024,cao_zero-shot_2024}. The first three tasks involve spatial degradation only: (1) \textbf{$4 \times$ super-resolution}, where low-resolution capture is simulated by applying $4\times$ average pooling to high-resolution videos; (2) \textbf{inpainting} with random masking at a missing rate of $r=0.5$; and (3) \textbf{motion deblurring}, where blurry videos are simulated by applying a $33 \times 33$ motion blur kernel of strength $0.5$ to clean videos. In addition, we examine (4) \textbf{temporal deconvolution}, where degraded videos are generated by applying a uniform point-spread-function (PSF) convolution of width $7$ along the temporal dimension, simulating the common artifact that multiple frames blend together in time-varying video capturing. The last task (5) \textbf{temporal deconvolution with spatial deblurring} combines (4) temporal deconvolution and (3) spatial motion deblurring, representing complex real-world scenarios. 

To generate evaluation datasets, we apply the above degradation models to four standard video benchmarks: DAVIS~\cite{perazzi_benchmark_2016}, REDS~\cite{nah_ntire_2019}, SPMCS~\cite{tao_detail-revealing_2017}, and UDM10~\cite{yi_progressive_2019}. To ensure consistent evaluation conditions, we process all videos by first segmenting them into sequential chunks of $8$ frames each, then center-cropping and resizing each frame to a uniform resolution of $512 \times 512$ pixels. 

We measure restoration quality using three standard spatial metrics: peak signal-to-noise ratio (PSNR), structural similarity index (SSIM), and learned perceptual image patch similarity (LPIPS, with the VGG backbone)~\cite{zhang_unreasonable_2018}. To assess temporal consistency, we employ warping error (WE)~\cite{lai_learning_2018}, a metric widely adopted in video processing research~\cite{zhou_upscale--video_2023,cao_zero-shot_2024,yeh_diffir2vr-zero_2024,zhou_propainter_2023,daras_warped_2024}. For all experiments, we compute OFs using RAFT~\cite{teed_raft_2020} and apply forward-backward consistency checking to generate non-occlusion masks for WE calculation.

\vspace{-1em}
\paragraph{\textcolor{umn_maroon}{Competing methods}} 
While our work focuses on zero-shot methods for VR using pre-trained image DMs, we benchmark our proposed method against both zero-shot and supervised methods. To ensure fair comparison, we select methods with publicly available implementations and evaluate them using their default settings wherever possible. For zero-shot methods, we compare against SVI~\cite{kwon_solving_2024}, VISION-XL (with SDXL)~\cite{kwon_vision-xl_2024}, VISION-base (with SD-base)~\cite{kwon_vision-xl_2024}, and DiffIR2VR (for super-resolution only)~\cite{yeh_diffir2vr-zero_2024}. For supervised methods, our benchmarks include the following: SD $\times 4$~\cite{rombach_high-resolution_2021}, VRT~\cite{liang_vrt_2022}, RealBasicVSR~\cite{chan_investigating_2021}, StableSR~\cite{wang_exploiting_2023}, and Upscale-A-Video (UAV)~\cite{zhou_upscale--video_2023} for super-resolution; SD Inpainting~\cite{rombach_high-resolution_2021} and ProPainter~\cite{zhou_propainter_2023} for inpainting; and VRT~\cite{liang_vrt_2022}, DeBlurGANv2~\cite{kupyn_deblurgan-v2_2019}, Stripformer~\cite{tsai_stripformer_2022}, and ID-Blau~\cite{Wu_IDBlau_2024} for motion deblurring and/or temporal deconvolution. \textbf{In all tables included in \cref{sec:results}, supervised and zero-shot methods are above and below the dotted line, respectively}.

\vspace{-1em}
\paragraph{\textcolor{umn_maroon}{Implementations details}} 
For all experiments except ablation studies, we use the pretrained Stable Diffusion (SD) 2.1-base model from~\cite{rombach_high-resolution_2021}~\footnote{\url{https://github.com/Stability-AI/stablediffusion}} and apply only $4$ reverse sampling steps ($T$) for $\mc R$ in our method. For text embeddings, we employ null-text following \cite{rout_solving_2023}. For the training objective, we employ a combination of \textit{MSE} and perceptual loss~\cite{zhang_unreasonable_2018} (with weight = $0.1$) as the loss function $\ell$ for both data fidelity (\cref{eq:ours4}) and progressive warping, except for inpainting tasks where we only use \textit{MSE} for data fidelity. We set the learning rates as $\num{5e-2}$ for the shared input seed $\mb z$ and $\num{1e-3}$ for the residuals. We use \textit{ADAM}~\cite{kingma_adam_2014} to solve the proposed formulation, which takes a total of $7$K iterations, with the progressive warping module activated after the initial $1.5$K iterations. To automate the activation of the progressive warping module, one could consider the quality of OF estimation and borrow ideas from zero-shot image restoration~\cite{li_self_2021,wang_early_2023}---a direction we leave for future work. The rank of the decomposed residual tensors is set to $k = 32$, empirically a good balance between computational efficiency and expressiveness. We leave the implementation details of the competing methods in \cref{app:implementation}. 


\subsection{Results}
\label{sec:results}
\begin{table}[!htpb]
    \centering
    \caption{(\textcolor{blue}{Spatial task}) Quantitative comparisons for video \textbf{super-resolution $\times 4$} (\textbf{Bold}: best, \underbar{under}: second best, \textcolor{darkpastelgreen}{green}: performance increase, \textcolor{red}{red}: performance decrease)}
    \resizebox{\columnwidth}{!}{ 
    \begin{tabular}{l c c c c c c c c}
        \toprule
        \multirow{2}{*}{Methods} & \multicolumn{4}{c}{DAVIS} & \multicolumn{4}{c}{REDS} \\
        \cmidrule(lr){2-5} \cmidrule(lr){6-9}
        & PSNR$\uparrow$ & SSIM$\uparrow$ & LPIPS$\downarrow$ & WE($10^{-2}$)$\downarrow$ & PSNR$\uparrow$ & SSIM$\uparrow$ & LPIPS$\downarrow$ & WE($10^{-2}$)$\downarrow$ \\
        \midrule
        SD $\times 4$~\cite{rombach_high-resolution_2021} & 24.33 & 0.615 & 0.358 & 1.523 & 23.57 & 0.619 & 0.371 & 1.769 \\
        VRT~\cite{liang_vrt_2022} & 25.97 & \underbar{0.780} & \underbar{0.295} & 1.063 & 24.54 & 0.756 & \underbar{0.305} & 1.266 \\
        RealBasicVSR~\cite{chan_investigating_2021} & 26.34 & 0.734 & \textbf{0.294} & \underbar{0.962} & \underbar{26.31} & \underbar{0.759} & \textbf{0.260} & \underbar{1.019} \\
        StableSR~\cite{wang_exploiting_2023} & 22.56 & 0.590 & 0.339 & 1.977 & 21.27 & 0.578 & 0.342 & 2.535 \\
        UAV~\cite{zhou_upscale--video_2023} & 23.65 & 0.589 & 0.397 & 1.623 & 23.02 & 0.593 & 0.422 & 1.920 \\
        \hdashline
        DiffIR2VR~\cite{yeh_diffir2vr-zero_2024} & 25.01 & 0.637 & 0.337 & 1.321 & 24.14 & 0.633 & 0.328 & 1.592 \\
        SVI~\cite{kwon_solving_2024} & 23.19 & 0.562 & 0.457 & 1.796 & 21.93 & 0.534 & 0.496 & 2.322 \\
        VISION-XL~\cite{kwon_vision-xl_2024} & \underbar{26.95} & 0.749 & 0.349 & 0.981 & 25.71 & 0.725 & 0.377 & 1.215 \\
        VISION-base~\cite{kwon_vision-xl_2024} & 26.17 & 0.712 & 0.338 & 1.137 & 24.95 & 0.687 & 0.376 & 1.406 \\
        \textbf{Ours} & \textbf{28.21} & \textbf{0.799} & 0.315 & \textbf{0.698} & \textbf{27.40} & \textbf{0.792} & 0.339 & \textbf{0.824} \\
        \textbf{Ours vs. Best compe.} & \textcolor{darkpastelgreen}{+1.26} & \textcolor{darkpastelgreen}{+0.019} & \textcolor{red}{+0.021} & \textcolor{darkpastelgreen}{-0.264} & \textcolor{darkpastelgreen}{+1.09} & \textcolor{darkpastelgreen}{+0.033} & \textcolor{red}{+0.079} & \textcolor{darkpastelgreen}{-0.195} \\
        \bottomrule
    \end{tabular}
    }
    
    \label{tab:sr_1}
\end{table}
\begin{table}[!htpb]
    \centering
    \caption{(\textcolor{blue}{Spatial task}) Quantitative comparisons for video \textbf{inpainting} with random masking $50 \%$ pixels (\textbf{Bold}: best, \underbar{under}: second best, \textcolor{darkpastelgreen}{green}: performance increase, \textcolor{red}{red}: performance decrease)}
    \resizebox{\columnwidth}{!}{ 
    \begin{tabular}{l c c c c c c c c}
        \toprule
        \multirow{2}{*}{Methods} & \multicolumn{4}{c}{DAVIS} & \multicolumn{4}{c}{REDS} \\
        \cmidrule(lr){2-5} \cmidrule(lr){6-9}
        & PSNR$\uparrow$ & SSIM$\uparrow$ & LPIPS$\downarrow$ & WE($10^{-2}$)$\downarrow$ & PSNR$\uparrow$ & SSIM$\uparrow$ & LPIPS$\downarrow$ & WE($10^{-2}$)$\downarrow$ \\
        \midrule
        SD Inpainting~\cite{rombach_high-resolution_2021} & 16.98 & 0.258 & 0.679 & 6.776 & 15.77 & 0.238 & 0.677 & 9.253 \\
        ProPainter~\cite{zhou_propainter_2023} & 28.60 & 0.823 & 0.281 & 0.655 & 28.05 & 0.827 & 0.273 & \underbar{0.733} \\
        \hdashline
        SVI~\cite{kwon_solving_2024} & 25.80 & 0.699 & 0.318 & 1.134 & 24.61 & 0.679 & 0.323 & 1.389 \\
        VISION-XL~\cite{kwon_vision-xl_2024} & \underbar{29.93} & \underbar{0.862} & \underbar{0.186} & \underbar{0.612} & \underbar{28.66} & \underbar{0.840} & \underbar{0.207} & 0.745 \\
        VISION-base~\cite{kwon_vision-xl_2024} & 26.54 & 0.732 & 0.286 & 1.033 & 25.30 & 0.711 & 0.293 & 1.268 \\
        \textbf{Ours} & \textbf{34.27} & \textbf{0.947} & \textbf{0.124} & \textbf{0.273} & \textbf{33.19} & \textbf{0.942} & \textbf{0.125} & \textbf{0.347} \\
        \textbf{Ours vs. Best compe.} & \textcolor{darkpastelgreen}{+4.34} & \textcolor{darkpastelgreen}{+0.085} & \textcolor{darkpastelgreen}{-0.062} & \textcolor{darkpastelgreen}{-0.339} & \textcolor{darkpastelgreen}{+4.53} & \textcolor{darkpastelgreen}{+0.102} & \textcolor{darkpastelgreen}{-0.082} & \textcolor{darkpastelgreen}{-0.386} \\
        \bottomrule
    \end{tabular}
    }
    
    \label{tab:inp_1}
\end{table}
\begin{table}[!htpb]
    \centering
    \caption{(\textcolor{blue}{Spatial task}) Quantitative comparisons for video \textbf{motion debluring} (\textbf{Bold}: best, \underbar{under}: second best, \textcolor{darkpastelgreen}{green}: performance increase, \textcolor{red}{red}: performance decrease)}
    \resizebox{\columnwidth}{!}{ 
    \begin{tabular}{l c c c c c c c c}
        \toprule
        \multirow{2}{*}{Methods} & \multicolumn{4}{c}{DAVIS} & \multicolumn{4}{c}{REDS} \\
        \cmidrule(lr){2-5} \cmidrule(lr){6-9}
        & PSNR$\uparrow$ & SSIM$\uparrow$ & LPIPS$\downarrow$ & WE($10^{-2}$)$\downarrow$ & PSNR$\uparrow$ & SSIM$\uparrow$ & LPIPS$\downarrow$ & WE($10^{-2}$)$\downarrow$ \\
        \midrule
        VRT~\cite{liang_vrt_2022} & 22.98 & 0.576 & 0.459 & 2.35 & 22.56 & 0.593 & 0.465 & 2.392 \\
        DeBlurGANv2~\cite{kupyn_deblurgan-v2_2019} & \underbar{24.32} & \underbar{0.649} & \underbar{0.371} & \underbar{1.734} & 24.11 & 0.677 & 0.368 & \underbar{1.722} \\
        Stripformer~\cite{tsai_stripformer_2022} & 24.07 & 0.612 & 0.381 & 2.218 & \underbar{24.72} & \underbar{0.699} & \underbar{0.343} & 1.797 \\
        ID-Blau~\cite{Wu_IDBlau_2024} & 23.07 & 0.589 & 0.397 & 2.623 & 23.84 & 0.654 & 0.359 & 2.198 \\
        \hdashline
        SVI~\cite{kwon_solving_2024} & 12.57 & 0.259 & 0.672 & 26.566 & 13.61 & 0.287 & 0.656 & 19.592 \\
        VISION-XL~\cite{kwon_vision-xl_2024} & 19.08 & 0.454 & 0.539 & 8.421 & 17.27 & 0.410 & 0.567 & 9.157 \\
        VISION-base~\cite{kwon_vision-xl_2024} & 12.81 & 0.290 & 0.699 & 26.359 & 13.98 & 0.320 & 0.684 & 21.882 \\
        \textbf{Ours} & \textbf{31.70} & \textbf{0.889} & \textbf{0.211} & \textbf{0.443} & \textbf{30.33} & \textbf{0.873} & \textbf{0.250} & \textbf{0.543} \\
        \textbf{Ours vs. Best compe.} & \textcolor{darkpastelgreen}{+7.38} & \textcolor{darkpastelgreen}{+0.240} & \textcolor{darkpastelgreen}{-0.160} & \textcolor{darkpastelgreen}{-1.291} & \textcolor{darkpastelgreen}{+5.61} & \textcolor{darkpastelgreen}{+0.174} & \textcolor{darkpastelgreen}{-0.093} & \textcolor{darkpastelgreen}{-1.179} \\
        \bottomrule
    \end{tabular}
    }
    
    \label{tab:deblur_1}
\end{table}
\begin{table}[!htpb]
    \centering
    \caption{(\textcolor{blue}{Temporal task}) Quantitative comparisons for video \textbf{temporal deconvolution} (\textbf{Bold}: best, \underbar{under}: second best, \textcolor{darkpastelgreen}{green}: performance increase, \textcolor{red}{red}: performance decrease)}
    \resizebox{\columnwidth}{!}{ 
    \begin{tabular}{l c c c c c c c c}
        \toprule
        \multirow{2}{*}{Methods} & \multicolumn{4}{c}{DAVIS} & \multicolumn{4}{c}{REDS} \\
        \cmidrule(lr){2-5} \cmidrule(lr){6-9}
        & PSNR$\uparrow$ & SSIM$\uparrow$ & LPIPS$\downarrow$ & WE($10^{-2}$)$\downarrow$ & PSNR$\uparrow$ & SSIM$\uparrow$ & LPIPS$\downarrow$ & WE($10^{-2}$)$\downarrow$ \\
        \midrule
        VRT~\cite{liang_vrt_2022} & 21.07 & 0.594 & 0.418 & 3.383 & 19.80 & 0.552 & 0.455 & 4.519 \\
        DeBlurGANv2~\cite{kupyn_deblurgan-v2_2019} & 20.89 & 0.586 & 0.414 & 3.514 & 19.58 & 0.544 & 0.446 & 4.734 \\
        Stripformer~\cite{tsai_stripformer_2022} & 20.75 & 0.565 & 0.411 & 3.638 & 19.49 & 0.524 & 0.453 & 4.893 \\
        ID-Blau~\cite{Wu_IDBlau_2024} & 18.42  & 0.491 & 0.476 & 6.128  &  17.73 & 0.467  & 0.515  & 6.968  \\   
        \hdashline
        SVI~\cite{kwon_solving_2024} & 27.63 & 0.766 & 0.151 & 0.901 & 26.14 & 0.740 & 0.162 & 1.119 \\
        VISION-XL~\cite{kwon_vision-xl_2024} & \underbar{31.79} & \underbar{0.908} & \underbar{0.125} & \underbar{0.501} &  \underbar{30.21} & \underbar{0.889} & \underbar{0.136} & \underbar{0.607} \\
        VISION-base~\cite{kwon_vision-xl_2024} & 27.66 & 0.767 & 0.151 & 0.897 & 26.17 & 0.741 & 0.162 & 1.115 \\
        \textbf{Ours} & \textbf{32.965} & \textbf{0.948} & \textbf{0.104} & \textbf{0.317} & \textbf{33.307} & \textbf{0.952} & \textbf{0.099} & \textbf{0.362} \\
        \textbf{Ours vs. Best compe.} & \textcolor{darkpastelgreen}{+1.17} & \textcolor{darkpastelgreen}{+0.040} & \textcolor{darkpastelgreen}{-0.021} & \textcolor{darkpastelgreen}{-0.184} & \textcolor{darkpastelgreen}{+3.09} & \textcolor{darkpastelgreen}{+0.063} & \textcolor{darkpastelgreen}{-0.037} & \textcolor{darkpastelgreen}{-0.245} \\
        \bottomrule
    \end{tabular}
    }
    
    \label{tab:temporal_1}
\end{table}
\begin{table}[!htpb]
    \centering
    \caption{(\textcolor{blue}{Spatio-temporal task}) Quantitative comparisons for video \textbf{temporal deconvolution with spatial deblurring} (\textbf{Bold}: best, \underbar{under}: second best, \textcolor{darkpastelgreen}{green}: performance increase, \textcolor{red}{red}: performance decrease)}
    \resizebox{\columnwidth}{!}{ 
    \begin{tabular}{l c c c c c c c c}
        \toprule
        \multirow{2}{*}{Methods} & \multicolumn{4}{c}{DAVIS} & \multicolumn{4}{c}{REDS} \\
        \cmidrule(lr){2-5} \cmidrule(lr){6-9}
        & PSNR$\uparrow$ & SSIM$\uparrow$ & LPIPS$\downarrow$ & WE($10^{-2}$)$\downarrow$ & PSNR$\uparrow$ & SSIM$\uparrow$ & LPIPS$\downarrow$ & WE($10^{-2}$)$\downarrow$ \\
        \midrule
        VRT~\cite{liang_vrt_2022} & 19.76 & 0.48 & 0.593 & 4.044 & 18.71 & 0.445 & 0.638 & 5.325 \\
        DeBlurGANv2~\cite{kupyn_deblurgan-v2_2019} & 19.91 & 0.496 & 0.556 & 3.921 & \underbar{18.89} & \underbar{0.460} & 0.600 & \underbar{5.219} \\
        Stripformer~\cite{tsai_stripformer_2022} & \underbar{20.00} & \underbar{0.503} & \underbar{0.548} & \underbar{3.879} & 18.88 & 0.458 & \underbar{0.588} & 5.224 \\
        ID-Blau~\cite{Wu_IDBlau_2024} & 19.76 & 0.489 & 0.554 & 4.060 & 18.73 & 0.450 & 0.593 & 5.418 \\   
        \hdashline
        SVI~\cite{kwon_solving_2024} & 12.21 & 0.260 & 0.705 & 27.955 & 12.27 & 0.255 & 0.711 & 28.575 \\
        VISION-XL~\cite{kwon_vision-xl_2024} & 15.94 & 0.346 & 0.627 & 15.220 & 16.70 & 0.393 & 0.637 & 12.675 \\
        VISION-base~\cite{kwon_vision-xl_2024} & 12.17 & 0.262 & 0.727 & 30.325 & 11.92 & 0.269 & 0.732 & 30.436 \\
        \textbf{Ours} & \textbf{26.97} & \textbf{0.778} & \textbf{0.343} & \textbf{0.852} & \textbf{26.61} & \textbf{0.763} & \textbf{0.381} & \textbf{0.982} \\
        \textbf{Ours vs. Best compe.} & \textcolor{darkpastelgreen}{+6.97} & \textcolor{darkpastelgreen}{+0.275} & \textcolor{darkpastelgreen}{-0.205} & \textcolor{darkpastelgreen}{-3.027} & \textcolor{darkpastelgreen}{+7.72} & \textcolor{darkpastelgreen}{+0.303} & \textcolor{darkpastelgreen}{-0.207} & \textcolor{darkpastelgreen}{-4.237} \\
        \bottomrule
    \end{tabular}
    }
    
    \label{tab:temporal_spatial_1}
\end{table}
As shown in \cref{tab:sr_1,tab:inp_1,tab:deblur_1,tab:temporal_1,tab:temporal_spatial_1}, our proposed method consistently outperforms existing ones across all VR tasks for almost all metrics on both the DAVIS and REDS datasets. We achieve consistent PSNR gains, ranging from 1--1.3dB in super-resolution, to the remarkable 6--8dB in combined degradation tasks, confirming our method's effectiveness across levels of degradation complexity. Most notably, our method delivers substantial improvements in temporal consistency, with Warping Error (WE) reductions ranging from 0.18--0.26 in simpler tasks and dramatic 3--4.2 in combined temporal deconvolution with spatial deblurring. Similar performance trends are also observed in the SPMCS and UDM10 datasets, with detailed results provided in \cref{app:more_res}. We also include qualitative comparisons in \cref{app:more_res}.


In particular, here, the test distribution may differ from the original training distribution for both supervised and zero-shot methods, potentially explaining the noticeable performance degradation compared to their originally reported results. Although we include supervised methods just for reference given their very different setting compared to the zero-shot approach we focus on, their relatively poor performance highlights their limited generalizability when confronted with real-world distribution shifts. For zero-shot competitors, our method consistently outperforms them by large margins in both spatial metrics and, most significantly, temporal consistency. These results validate the effectiveness of our proposed formulation in \cref{eq:ours4} and our hierarchical framework for bilevel temporal consistency described in \cref{subsec:temporal}. 

A particularly interesting observation is that CG-based methods~\cite{kwon_solving_2024,kwon_vision-xl_2024} perform \emph{uniformly} poorly on motion deblur-related tasks, as is evident in \cref{tab:deblur_1,tab:temporal_spatial_1}. This observation is consistent with our theoretical analysis in \cref{sec:bg}, which highlighted that \textit{CG requires the degradation operator $\mathcal{A}$ be known, symmetric and positive definite}~\cite{shewchuk_introduction_1994}. When CG deals with VR tasks that violate these mathematical requirements---such as motion deblurring---numerical issues arise, leading to poor performance. This limitation comprises a critical weakness in the SOTA CG-based methods, whereas our method is versatile and remains effective for different degradation scenarios.

\subsection{Ablation studies}
\label{sec:ablation}


\paragraph{\textcolor{umn_maroon}{Effects of noise prior and progressive warping}} 
\cref{tab:ablation_stages} presents the quantitative results of our ablation studies on the DAVIS dataset for video super-resolution. The integration of noise prior significantly improves the baseline model performance in terms of both PSNR and SSIM, while substantially reducing WE, signifying enhanced semantic-level temporal consistency. Progressive warping yields even stronger results with marked improvements in all metrics compared to the baseline, particularly in WE, indicating superior pixel-level temporal consistency. Notably, combining both components produces the optimal configuration, with our complete model outperforming an SOTA method~\cite{kwon_vision-xl_2024} in all metrics. These results confirm that the noise prior and progressive warping mechanisms complement each other in promoting bilevel temporal consistency. 

\vspace{-1em}
\paragraph{\textcolor{umn_maroon}{Effect of controllable residuals}} 
We also study how the learning rate for residuals (LR$_r$) and the radius of residual balls (ie, controlling the magnitude of residuals) affect the performance. Intuitively, the residuals should not be too large to allow substantial deviation from the image manifold, and not be too small to limit their powers in modeling reasonable framewise deviation from the shared frame. This intuition is confirmed by our data in \cref{tab:ablation_lr_radius}: a restrictive radius (0.1) prevents adequate motion learning, whereas a moderate radius (1.0) allows effective discrepancy modeling; similarly, learning rate calibration is critical---a high rate (0.01) causes residual learning to dominate and downplays the influence from the DM prior, while an insufficient rate (0.0001) limits framewise adaptation. The optimal configuration (LR$_r$=0.001, radius=1.0) achieves the best performance by balancing these competing factors. 

\begin{table}[!htpb]
    \centering
    \caption{Ablation study on two essential components for bilevel temporal consistency, performed on DAVIS dataset for video super-resolution $\times 4$. (\textbf{Bold}: best, \underline{under}: second best)}
    \resizebox{0.85\columnwidth}{!}{ 
    \begin{tabular}{c c c c c}
        \toprule
        Method & PSNR$\uparrow$ & SSIM$\uparrow$ & LPIPS$\downarrow$ & WE($10^{-2}$)$\downarrow$ \\
        \midrule
        SOTA \cite{kwon_vision-xl_2024} & 26.03 & 0.717 & 0.339 & 1.411 \\
        \hdashline
        Base & 24.70 & 0.612 & 0.366 & 1.398 \\
        Base with noise prior & 26.09 & 0.703 & 0.410 & 1.057 \\
        Base with warping & \underline{27.14} & \underline{0.736} & \textbf{0.301} & \underline{0.943} \\
        Base with both& \textbf{27.95} & \textbf{0.790} & \underline{0.321} & \textbf{0.725} \\
        \bottomrule
    \end{tabular}
    }
    
    \label{tab:ablation_stages}
\end{table}

\begin{table}[!htpb]
    \centering
    \caption{Ablation study on the trainable residuals, performed on DAVIS dataset for video super-resolution $\times 4$.}
    \resizebox{0.85\columnwidth}{!}{ 
    \begin{tabular}{c c c c c c}
        \toprule
        LR$_r$ & Radius & PSNR$\uparrow$ & SSIM$\uparrow$ & LPIPS$\downarrow$ & WE($10^{-2}$)$\downarrow$ \\
        \midrule
        \multirow{3}{*}{0.01} & 0.1 & 24.36 & 0.654 & 0.436 & 1.557 \\
                            & 1.0 & 27.19 & 0.741 & 0.402 & 0.804 \\
                            & 10.0 & 27.16 & 0.739 & 0.405 & 0.806 \\
        \midrule
        \multirow{3}{*}{0.001} & 0.1 & 25.74 & 0.719 & 0.374 & 1.161 \\
                            & 1.0 & \textbf{27.95} & \textbf{0.790} & \textbf{0.321} & \textbf{0.725} \\
                            & 10.0 & {27.91} & {0.789} & {0.324} & {0.737} \\
        \midrule
        \multirow{3}{*}{0.0001} & 0.1 & 25.11 & 0.677 & 0.389 & 1.298 \\
                            & 1.0 & 26.17 & 0.709 & 0.366 & 1.096 \\
                            & 10.0 & 26.17 & 0.710 & 0.369 & 1.089 \\
        \bottomrule
    \end{tabular}
    }
    \label{tab:ablation_lr_radius}
\end{table}
\section{Discussion}
\label{sec:dis}

In this paper, we focus on solving video restoration (VR) problems using pre-trained image DMs. We systematically address three key challenges: approximation errors through a novel MAP framework that reparameterizes frames directly in the seed space of LDMs; temporal inconsistency via a hierarchical bilevel consistency strategy; and high computational and memory demands through reformulation and low-rank decomposition. Comprehensive experiments show that our method significantly improves over state-of-the-art methods across various VR tasks without task-specific training, in terms of both frame quality and temporal consistency. 

As for limitations, our empirical results in \cref{tab:ablation_steps} suggest a fundamental gap between image generation and regression using pretrained DMs, while \cref{fig:cluster} shows the seed space clustering phenomenon we have not fully explained. Our work remains primarily empirical, and we leave a solid theoretical understanding for future research.

\section*{Acknowledgment}
Part of the work was done when Wang H. was interning at Amazon.com, Inc. during the summer of 2024. Wang H. and Sun J. are also partially supported by a UMN DSI Seed Grant. The authors acknowledge the Minnesota Supercomputing Institute (MSI) at the University of Minnesota for providing resources that contributed to the research results reported in this article. 


{
    \small
    \bibliographystyle{IEEEtran}
    \bibliography{main}
}



\newpage  
\appendix

\onecolumn


\section{More implementation details}
\label{app:implementation}

\paragraph{\textcolor{umn_maroon}{Implementations of competing methods}} 
We implement most competing methods using their default code and settings. For SVI, we adapt the implementation from the VISION-XL codebase by deactivating both the low-pass filter and DDIM inversion initialization. To ensure a fair comparison, we use the Stable Diffusion 2.1-base as the backbone for SVI. Similarly, for VISION-base, we modify the VISION-XL code and also take Stable Diffusion 2.1-base as the backbone. When applying SVI, VISION-XL, and VISION-base to motion-deblurring tasks, we implement an additional stopping mechanism that is activated just before the numerical issue due to conjugate gradient (CG) arises. This modification is necessary because, as discussed in \cref{sec:bg}, CG requires the degradation operator $\mathcal{A}$ to be known, symmetric, positive definite to have guaranteed convergence and be free from numerical problems~\cite{shewchuk_introduction_1994}. Motion deblurring operations inherently violate these CG requirements, as the degradation operators involved are asymmetric and tend to be ill-conditioned. The official implementations used for all competitors are listed below. 

 {\small
\begin{itemize} 
    \item SVI~\cite{kwon_solving_2024}: \url{https://github.com/vision-xl/codes}
    \item VISION-XL (with SDXL)~\cite{kwon_vision-xl_2024}: \url{https://github.com/vision-xl/codes}
    \item VISION-base (with SD-base)~\cite{kwon_vision-xl_2024}: \url{https://github.com/vision-xl/codes}
    \item DiffIR2VR~\cite{yeh_diffir2vr-zero_2024}: \url{https://github.com/jimmycv07/DiffIR2VR-Zero}
    \item SD $\times 4$~\cite{rombach_high-resolution_2021}: \url{https://github.com/Stability-AI/stablediffusion}
    \item VRT~\cite{liang_vrt_2022}: \url{https://github.com/JingyunLiang/VRT}
    \item RealBasicVSR~\cite{chan_investigating_2021}: \url{https://github.com/ckkelvinchan/RealBasicVSR}
    \item StableSR~\cite{wang_exploiting_2023}: \url{https://github.com/IceClear/StableSR}
    \item Upscale-A-Video (UAV)~\cite{zhou_upscale--video_2023}: \url{https://github.com/sczhou/Upscale-A-Video}
    \item SD Inpainting~\cite{rombach_high-resolution_2021}: \url{https://github.com/Stability-AI/stablediffusion}
    \item ProPainter~\cite{zhou_propainter_2023}: \url{https://github.com/sczhou/ProPainter}
    \item DeBlurGANv2~\cite{kupyn_deblurgan-v2_2019}: \url{https://github.com/VITA-Group/DeblurGANv2}
    \item Stripformer~\cite{tsai_stripformer_2022}: \url{https://github.com/pp00704831/Stripformer-ECCV-2022-}
    \item ID-Blau~\cite{Wu_IDBlau_2024}: \url{https://github.com/plusgood-steven/ID-Blau}
\end{itemize} 
 }

\section{More experiment results}
\label{app:more_res}

\subsection{Quantitative results}

We provide more quantitative results for the five video restoration problems, including super-resolution, inpainting, motion deblurring, temporal deconvolution, and temporal deconvolution with spatial deblurring on SPMCS and UDM10 in \cref{tab:sr_2,tab:inp_2,tab:deblur_2,tab:temporal_2,tab:temporal_spatial_2}.

\begin{table}[!htpb]
    \centering
    \caption{(\textcolor{blue}{Spatial task}) Quantitative comparisons for video \textbf{super-resolution $\times 4$} (\textbf{Bold}: best, \underbar{under}: second best, \textcolor{darkpastelgreen}{green}: performance increase, \textcolor{red}{red}: performance decrease)}
    \resizebox{0.9\columnwidth}{!}{ 
    \begin{tabular}{l c c c c c c c c}
        \toprule
        \multirow{2}{*}{Methods} & \multicolumn{4}{c}{SPMCS} & \multicolumn{4}{c}{UDM10} \\
        \cmidrule(lr){2-5} \cmidrule(lr){6-9}
        & PSNR$\uparrow$ & SSIM$\uparrow$ & LPIPS$\downarrow$ & WE($10^{-2}$)$\downarrow$ & PSNR$\uparrow$ & SSIM$\uparrow$ & LPIPS$\downarrow$ & WE($10^{-2}$)$\downarrow$ \\
        \midrule
        SD $\times 4$~\cite{rombach_high-resolution_2021} & 23.71 & 0.631 & 0.365 & 2.094 & 26.57 & 0.749 & 0.323 & 0.921 \\
        VRT~\cite{liang_vrt_2022} & 26.19 & \textbf{0.811} & \textbf{0.243} & \underbar{1.169} & 28.11 & \underbar{0.865} & \textbf{0.179} & \underbar{0.609} \\
        RealBasicVSR~\cite{chan_investigating_2021} & 25.77 & 0.746 & \underbar{0.288} & 1.363 & 28.27 & 0.845 & \underbar{0.245} & 0.640 \\
        StableSR~\cite{wang_exploiting_2023} & 21.82 & 0.598 & 0.352 & 2.908 & 23.11 & 0.72 & 0.319 & 1.76 \\
        UAV~\cite{zhou_upscale--video_2023} & 23.05 & 0.611 & 0.383 & 2.351 & 25.59 & 0.750 & 0.331 & 1.049 \\
        \hdashline
        DiffIR2VR~\cite{yeh_diffir2vr-zero_2024} & 24.61 & 0.649 & 0.347 & 1.906 & 27.41 & 0.793 & 0.290 & 0.824 \\
        SVI~\cite{kwon_solving_2024} & 22.58 & 0.581 & 0.426 & 2.469 & 23.80 & 0.644 & 0.440 & 1.596 \\
        VISION-XL~\cite{kwon_vision-xl_2024} & \underbar{26.43} & 0.758 & 0.314 & 1.310 & \underbar{29.22} & 0.847 & 0.268 & 0.616 \\
        VISION-base~\cite{kwon_vision-xl_2024} & 25.74 & 0.718 & 0.321 & 1.572 & 28.63 & 0.828 & 0.247 & 0.713 \\
        \textbf{Ours} & \textbf{27.40} & \underbar{0.793} & 0.298 & \textbf{1.073} & \textbf{31.32} & \textbf{0.886} & 0.254 & \textbf{0.413} \\
        \textbf{Ours vs. Best compe.} & \textcolor{darkpastelgreen}{+0.97} & \textcolor{red}{-0.018} & \textcolor{red}{+0.055} & \textcolor{darkpastelgreen}{-0.096} & \textcolor{darkpastelgreen}{+2.10} & \textcolor{darkpastelgreen}{+0.021} & \textcolor{red}{+0.075} & \textcolor{darkpastelgreen}{-0.196} \\
        \bottomrule
    \end{tabular}
    }
    
    \label{tab:sr_2}
\end{table}

\begin{table}[!htpb]
    \centering
    \caption{(\textcolor{blue}{Spatial task}) Quantitative comparisons for video \textbf{inpainting} with random masking $50 \%$ pixels (\textbf{Bold}: best, \underbar{under}: second best, \textcolor{darkpastelgreen}{green}: performance increase, \textcolor{red}{red}: performance decrease)}
    \resizebox{0.9\columnwidth}{!}{ 
    \begin{tabular}{l c c c c c c c c}
        \toprule
        \multirow{2}{*}{Methods} & \multicolumn{4}{c}{SPMCS} & \multicolumn{4}{c}{UDM10} \\
        \cmidrule(lr){2-5} \cmidrule(lr){6-9}
        & PSNR$\uparrow$ & SSIM$\uparrow$ & LPIPS$\downarrow$ & WE($10^{-2}$)$\downarrow$ & PSNR$\uparrow$ & SSIM$\uparrow$ & LPIPS$\downarrow$ & WE($10^{-2}$)$\downarrow$ \\
        \midrule
        SD Inpainting~\cite{rombach_high-resolution_2021} & 16.62 & 0.276 & 0.709 & 8.969 & 15.28 & 0.243 & 0.758 & 10.533 \\
        ProPainter~\cite{zhou_propainter_2023} & 28.61 & 0.854 & 0.290 & 0.618 & 30.36 & 0.855 & 0.317 & 0.412 \\
        \hdashline
        SVI~\cite{kwon_solving_2024} & 25.92 & 0.746 & 0.334 & 1.360 & 28.05 & 0.817 & 0.323 & 0.658 \\
        VISION-XL~\cite{kwon_vision-xl_2024} & \underbar{30.14} & \underbar{0.894} & \underbar{0.178} & \underbar{0.527} & \underbar{32.04} & \underbar{0.906} & \underbar{0.199} & \underbar{0.345} \\
        VISION-base~\cite{kwon_vision-xl_2024} & 26.75 & 0.778 & 0.290 & 1.212 & 29.51 & 0.857 & 0.242 & 0.553 \\
        \textbf{Ours} & \textbf{36.92} & \textbf{0.968} & \textbf{0.083} & \textbf{0.174} & \textbf{38.75} & \textbf{0.974} & \textbf{0.087} & \textbf{0.141} \\
        \textbf{Ours vs. Best compe.} & \textcolor{darkpastelgreen}{+6.78} & \textcolor{darkpastelgreen}{+0.074} & \textcolor{darkpastelgreen}{-0.095} & \textcolor{darkpastelgreen}{-0.353} & \textcolor{darkpastelgreen}{+6.71} & \textcolor{darkpastelgreen}{+0.068} & \textcolor{darkpastelgreen}{-0.112} & \textcolor{darkpastelgreen}{-0.204} \\
        \bottomrule
    \end{tabular}
    }
    
    \label{tab:inp_2}
\end{table}

\begin{table}[!htpb]
    \centering
    \caption{(\textcolor{blue}{Spatial task}) Quantitative comparisons for video \textbf{motion debluring} (\textbf{Bold}: best, \underbar{under}: second best, \textcolor{darkpastelgreen}{green}: performance increase, \textcolor{red}{red}: performance decrease)}
    \resizebox{0.9\columnwidth}{!}{ 
    \begin{tabular}{l c c c c c c c c}
        \toprule
        \multirow{2}{*}{Methods} & \multicolumn{4}{c}{SPMCS} & \multicolumn{4}{c}{UDM10} \\
        \cmidrule(lr){2-5} \cmidrule(lr){6-9}
        & PSNR$\uparrow$ & SSIM$\uparrow$ & LPIPS$\downarrow$ & WE($10^{-2}$)$\downarrow$ & PSNR$\uparrow$ & SSIM$\uparrow$ & LPIPS$\downarrow$ & WE($10^{-2}$)$\downarrow$ \\
        \midrule
        VRT~\cite{liang_vrt_2022} & 22.85 & 0.572 & 0.431 & 3.133 & 23.38 & 0.694 & 0.39 & 2.315 \\
        DeBlurGANv2~\cite{kupyn_deblurgan-v2_2019} & \underbar{24.47} & \underbar{0.674} & 0.338 & \underbar{2.181} & \underbar{25.40} & \underbar{0.776} & 0.299 & \underbar{1.465} \\
        Stripformer~\cite{tsai_stripformer_2022} & 24.41 & 0.646 & \underbar{0.334} & 2.736 & 25.23 & 0.758 & \underbar{0.296} & 1.842 \\
        ID-Blau~\cite{Wu_IDBlau_2024} & 23.91 & 0.623 & 0.336 & 2.962 & 24.47 & 0.719 & 0.315 & 2.307 \\
        \hdashline
        SVI~\cite{kwon_solving_2024} & 14.12 & 0.310 & 0.645 & 24.896 & 14.66 & 0.369 & 0.677 & 20.061 \\
        VISION-XL~\cite{kwon_vision-xl_2024} & 18.89 & 0.452 & 0.514 & 12.202 & 19.87 & 0.559 & 0.495 & 5.560 \\
        VISION-base~\cite{kwon_vision-xl_2024} & 13.39 & 0.296 & 0.676 & 27.699 & 12.04 & 0.300 & 0.734 & 32.348 \\
        \textbf{Ours} & \textbf{33.47} & \textbf{0.923} & \textbf{0.166} & \textbf{0.314} & \textbf{34.32} & \textbf{0.920} & \textbf{0.207} & \textbf{0.259} \\
        \textbf{Ours vs. Best compe.} & \textcolor{darkpastelgreen}{+9.00} & \textcolor{darkpastelgreen}{+0.249} & \textcolor{darkpastelgreen}{-0.168} & \textcolor{darkpastelgreen}{-1.867} & \textcolor{darkpastelgreen}{+8.92} & \textcolor{darkpastelgreen}{+0.144} & \textcolor{darkpastelgreen}{-0.089} & \textcolor{darkpastelgreen}{-1.206} \\
        \bottomrule
    \end{tabular}
    }
    
    \label{tab:deblur_2}
\end{table}

\begin{table}[!htpb]
    \centering
    \caption{(\textcolor{blue}{Temporal task}) Quantitative comparisons for video \textbf{temporal deconvolution} (\textbf{Bold}: best, \underbar{under}: second best, \textcolor{darkpastelgreen}{green}: performance increase, \textcolor{red}{red}: performance decrease)}
    \resizebox{0.9\columnwidth}{!}{ 
    \begin{tabular}{l c c c c c c c c}
        \toprule
        \multirow{2}{*}{Methods} & \multicolumn{4}{c}{SPMCS} & \multicolumn{4}{c}{UDM10} \\
        \cmidrule(lr){2-5} \cmidrule(lr){6-9}
        & PSNR$\uparrow$ & SSIM$\uparrow$ & LPIPS$\downarrow$ & WE($10^{-2}$)$\downarrow$ & PSNR$\uparrow$ & SSIM$\uparrow$ & LPIPS$\downarrow$ & WE($10^{-2}$)$\downarrow$ \\
        \midrule
        VRT~\cite{liang_vrt_2022} & 31.65 & 0.867 & 0.153 & 0.888 & 24.53 & 0.762 & 0.318 & 1.645 \\
        DeBlurGANv2~\cite{kupyn_deblurgan-v2_2019} & 30.79 & 0.856 & 0.136 & 0.991 & 24.49 & 0.753 & 0.284 & 1.751 \\
        Stripformer~\cite{tsai_stripformer_2022} & 31.38 & 0.861 & 0.121 & 0.969 & 24.81 & 0.757 & 0.247 & 1.798 \\
        ID-Blau~\cite{Wu_IDBlau_2024} & 29.37 & 0.838 & 0.140 & 1.290 & 24.38 & 0.753 & 0.256 & 1.973 \\
        \hdashline
        SVI~\cite{kwon_solving_2024} & 28.18 & 0.821 & 0.135 & 0.958 & 31.41 & 0.891 & 0.105 & 0.414 \\
        VISION-XL~\cite{kwon_vision-xl_2024} & \underbar{32.10} & \underbar{0.931} & \underbar{0.102} & \underbar{0.383} & \underbar{34.48} & \underbar{0.945} & \textbf{0.076} & \underbar{0.256} \\
        VISION-base~\cite{kwon_vision-xl_2024} & 28.21 & 0.821 & 0.135 & 0.954 & 31.44 & 0.891 & 0.104 & 0.412 \\
        \textbf{Ours} & \textbf{40.45} & \textbf{0.987} & \textbf{0.030} & \textbf{0.095} & \textbf{36.67} & \textbf{0.971} & \underbar{0.079} & \textbf{0.180} \\
        \textbf{Ours vs. Best compe.} & \textcolor{darkpastelgreen}{+8.35} & \textcolor{darkpastelgreen}{+0.056} & \textcolor{darkpastelgreen}{-0.072} & \textcolor{darkpastelgreen}{-0.288} & \textcolor{darkpastelgreen}{+2.19} & \textcolor{darkpastelgreen}{+0.026} & \textcolor{red}{+0.003} & \textcolor{darkpastelgreen}{-0.076} \\
        \bottomrule
    \end{tabular}
    }
    
    \label{tab:temporal_2}
\end{table}

\begin{table}[!htpb]
    \centering
    \caption{(\textcolor{blue}{Spatio-temporal task}) Quantitative comparisons for video \textbf{temporal deconvolution with spatial deblurring} (\textbf{Bold}: best, \underbar{under}: second best, \textcolor{darkpastelgreen}{green}: performance increase, \textcolor{red}{red}: performance decrease)}
    \resizebox{0.9\columnwidth}{!}{ 
    \begin{tabular}{l c c c c c c c c}
        \toprule
        \multirow{2}{*}{Methods} & \multicolumn{4}{c}{SPMCS} & \multicolumn{4}{c}{UDM10} \\
        \cmidrule(lr){2-5} \cmidrule(lr){6-9}
        & PSNR$\uparrow$ & SSIM$\uparrow$ & LPIPS$\downarrow$ & WE($10^{-2}$)$\downarrow$ & PSNR$\uparrow$ & SSIM$\uparrow$ & LPIPS$\downarrow$ & WE($10^{-2}$)$\downarrow$ \\
        \midrule
        VRT~\cite{liang_vrt_2022} & 22.14 & 0.548 & 0.488 & 3.343 & 22.15 & 0.661 & 0.472 & 2.594 \\
        DeBlurGANv2~\cite{kupyn_deblurgan-v2_2019} & 23.32 & \underbar{0.638} & 0.383 & \underbar{2.498} & \underbar{22.83} & 0.685 & 0.426 & \underbar{2.257} \\
        Stripformer~\cite{tsai_stripformer_2022} & \underbar{23.40} & 0.621 & 0.382 & 2.898 & 22.79 & \underbar{0.688} & 0.423 & 2.361 \\
        ID-Blau~\cite{Wu_IDBlau_2024} & 23.14 & 0.609 & \underbar{0.379} & 3.077 & 22.72 & 0.682 & \underbar{0.422} & 2.459 \\   
        \hdashline
        SVI~\cite{kwon_solving_2024} & 12.70 & 0.273 & 0.670 & 29.942 & 13.66 & 0.363 & 0.693 & 23.061 \\
        VISION-XL~\cite{kwon_vision-xl_2024} & 18.39 & 0.460 & 0.535 & 10.862 & 17.68 & 0.500 & 0.5463 & 9.576 \\
        VISION-base~\cite{kwon_vision-xl_2024} & 12.84 & 0.291 & 0.694 & 28.502 & 12.40 & 0.334 & 0.742 & 33.661 \\
        \textbf{Ours} & \textbf{30.87} & \textbf{0.886} & \textbf{0.215} & \textbf{0.535} & \textbf{29.88} & \textbf{0.857} & \textbf{0.292} & \textbf{0.534} \\
        \textbf{Ours vs. Best compe.} & \textcolor{darkpastelgreen}{+7.47} & \textcolor{darkpastelgreen}{+0.248} & \textcolor{darkpastelgreen}{-0.164} & \textcolor{darkpastelgreen}{-1.963} & \textcolor{darkpastelgreen}{+7.05} & \textcolor{darkpastelgreen}{+0.169} & \textcolor{darkpastelgreen}{-0.130} & \textcolor{darkpastelgreen}{-1.723} \\
        \bottomrule
    \end{tabular}
    }
    
    \label{tab:temporal_spatial_2}
\end{table}

\subsection{Qualitative results}

\begin{figure}[!htpb]
    \centering
    \includegraphics[width=0.9\textwidth]{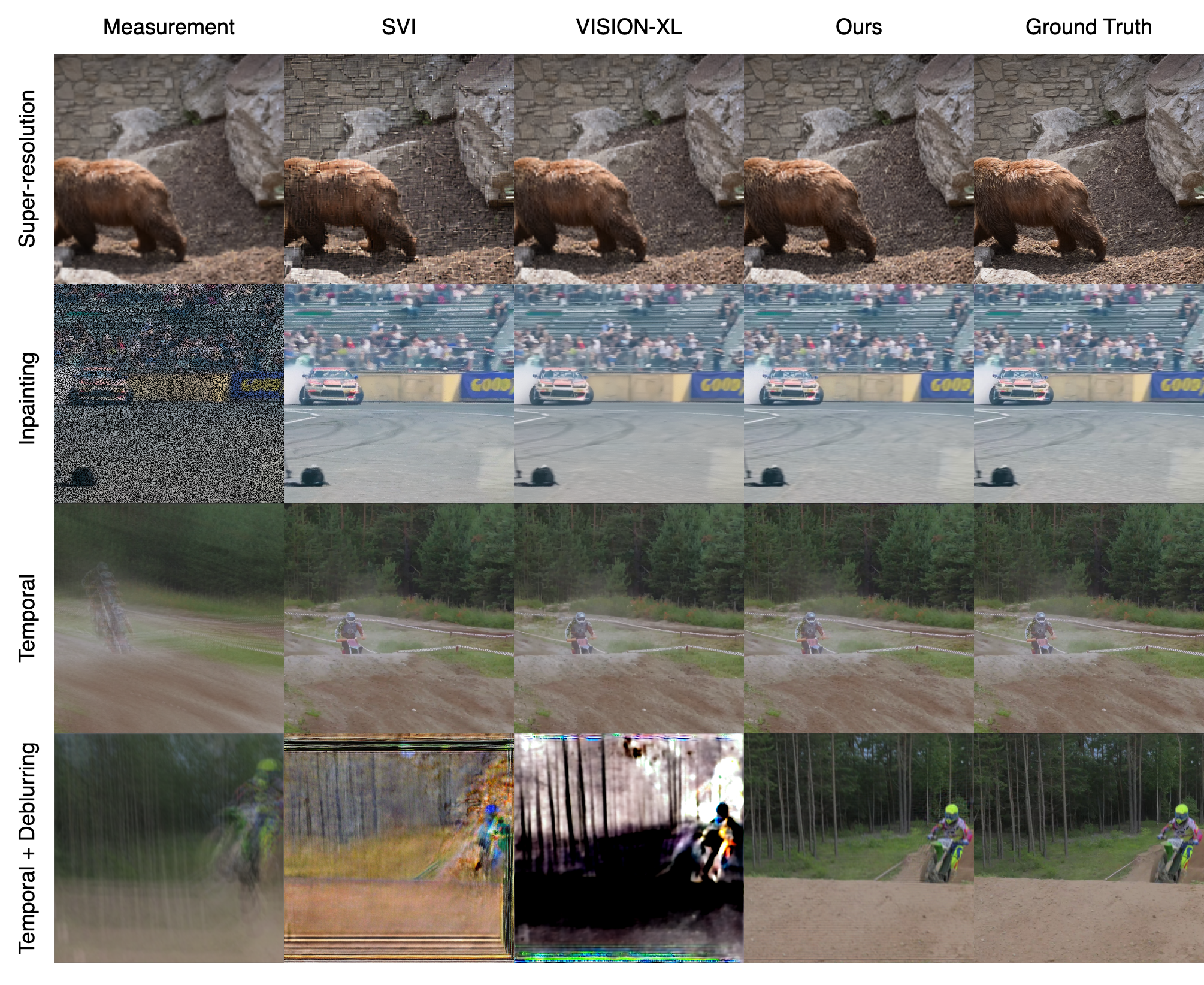}
    \caption{Qualitative comparisons on the DAVIS dataset: (a) Super-resolution $\times 4$, (b) Inpainting with $50\%$ random pixel masking, (c) Temporal deconvolution using uniform PSF with kernel width $k = 7$, and (d) Temporal deconvolution with spatial deblurring.}
    \label{fig:comparisons}
\end{figure}
\begin{figure*}
    \begin{minipage}{0.01\textwidth}
        \rotatebox{90}{Measurement}
    \end{minipage}%
    \begin{minipage}{0.99\textwidth}
        \centering
        \includegraphics[width=0.96\textwidth]{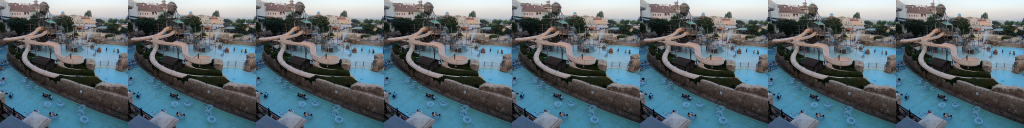}
    \end{minipage}
    
    \vspace{0.3cm}

    \begin{minipage}{0.01\textwidth}
        \rotatebox{90}{RealBasicVSR}
    \end{minipage}%
    \begin{minipage}{0.99\textwidth}
        \centering
        \includegraphics[width=0.96\textwidth]{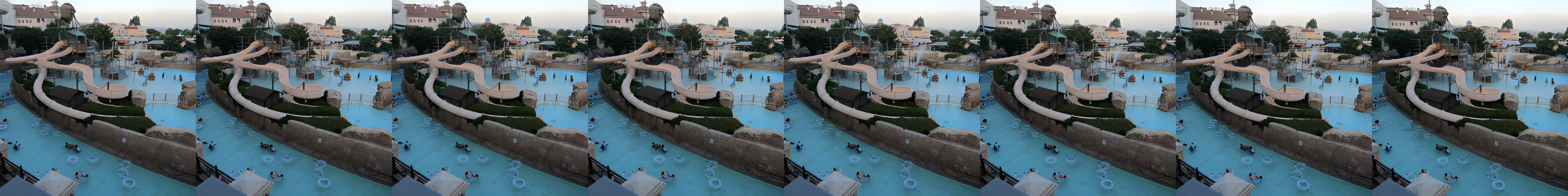}
    \end{minipage}
    
    \vspace{0.3cm}

    \begin{minipage}{0.01\textwidth}
        \rotatebox{90}{SVI}
    \end{minipage}%
    \begin{minipage}{0.99\textwidth}
        \centering
        \includegraphics[width=0.96\textwidth]{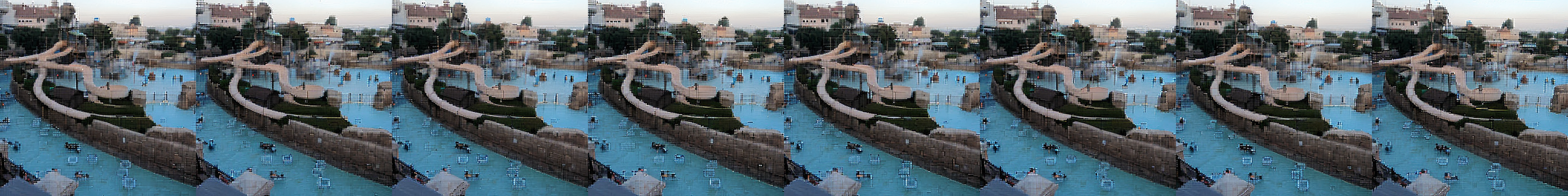}
    \end{minipage}
    
    \vspace{0.3cm}

    \begin{minipage}{0.01\textwidth}
        \rotatebox{90}{VISION-XL}
    \end{minipage}%
    \begin{minipage}{0.99\textwidth}
        \centering
        \includegraphics[width=0.96\textwidth]{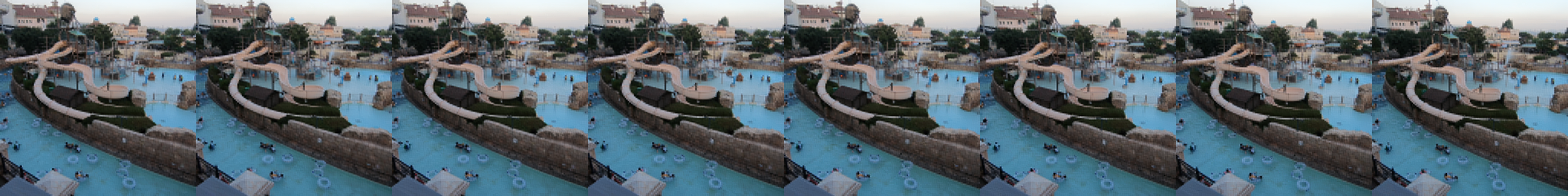}
    \end{minipage}
    
    \vspace{0.3cm}

    \begin{minipage}{0.01\textwidth}
        \rotatebox{90}{Ours}
    \end{minipage}%
    \begin{minipage}{0.99\textwidth}
        \centering
        \includegraphics[width=0.96\textwidth]{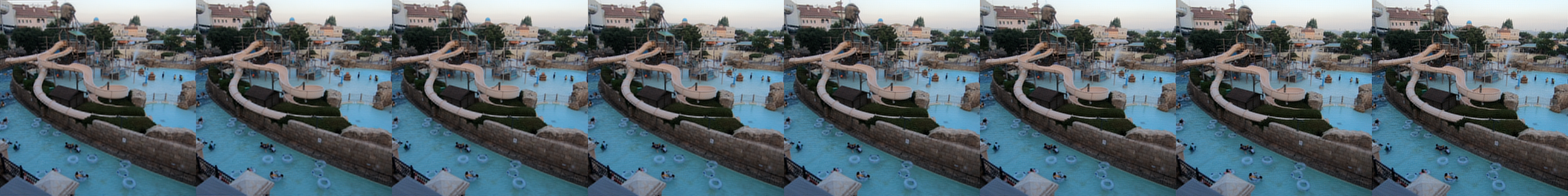}
    \end{minipage}
    
    \vspace{0.3cm}
    
   \begin{minipage}{0.01\textwidth}
        \rotatebox{90}{Ground Truth}
    \end{minipage}%
    \begin{minipage}{0.99\textwidth}
        \centering
        \includegraphics[width=0.96\textwidth]{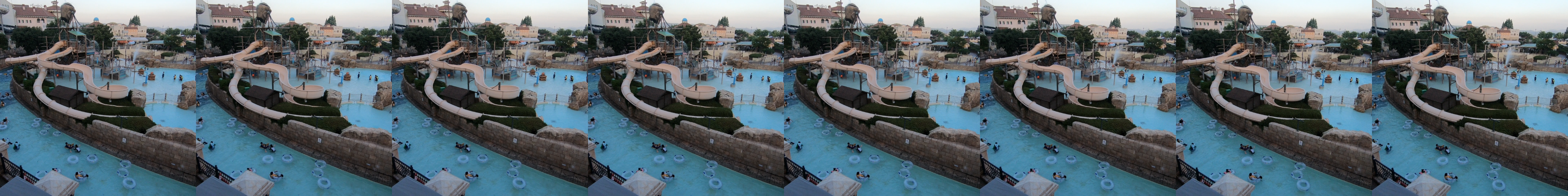}
    \end{minipage}
    \caption{Qualitative comparisons for video \textbf{super-resolution $\times 4$}}
    \label{fig:video_comparison_sr}
\end{figure*}
\begin{figure*}
    \begin{minipage}{0.01\textwidth}
        \rotatebox{90}{Measurement}
    \end{minipage}%
    \begin{minipage}{0.99\textwidth}
        \centering
        \includegraphics[width=0.96\textwidth]{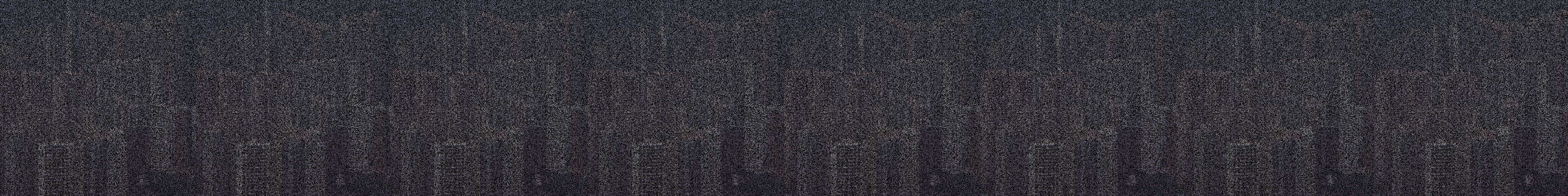}
    \end{minipage}
    
    \vspace{0.3cm}

    \begin{minipage}{0.01\textwidth}
        \rotatebox{90}{Pro Painter}
    \end{minipage}%
    \begin{minipage}{0.99\textwidth}
        \centering
        \includegraphics[width=0.96\textwidth]{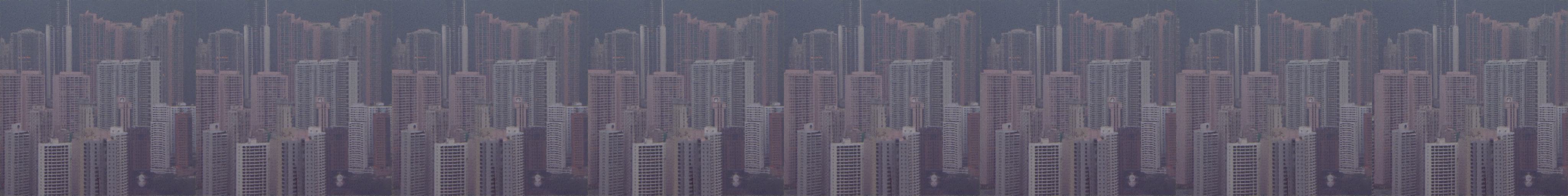}
    \end{minipage}
    
    \vspace{0.3cm}

    \begin{minipage}{0.01\textwidth}
        \rotatebox{90}{SVI}
    \end{minipage}%
    \begin{minipage}{0.99\textwidth}
        \centering
        \includegraphics[width=0.96\textwidth]{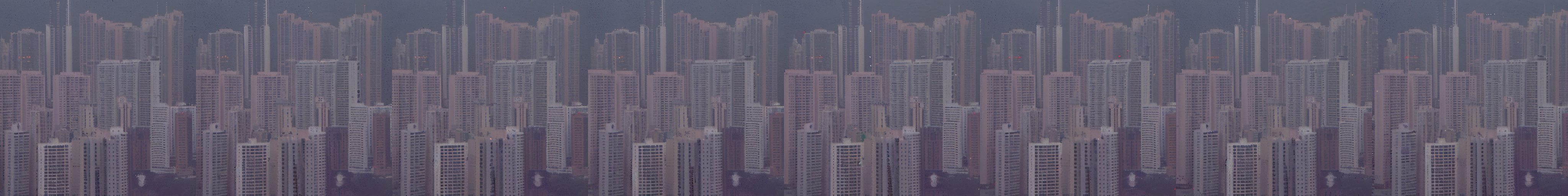}
    \end{minipage}
    
    \vspace{0.3cm}

    \begin{minipage}{0.01\textwidth}
        \rotatebox{90}{VISION-XL}
    \end{minipage}%
    \begin{minipage}{0.99\textwidth}
        \centering
        \includegraphics[width=0.96\textwidth]{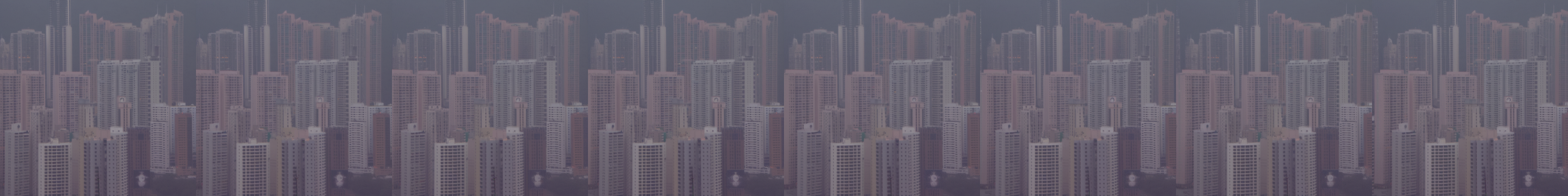}
    \end{minipage}
    
    \vspace{0.3cm}

    \begin{minipage}{0.01\textwidth}
        \rotatebox{90}{Ours}
    \end{minipage}%
    \begin{minipage}{0.99\textwidth}
        \centering
        \includegraphics[width=0.96\textwidth]{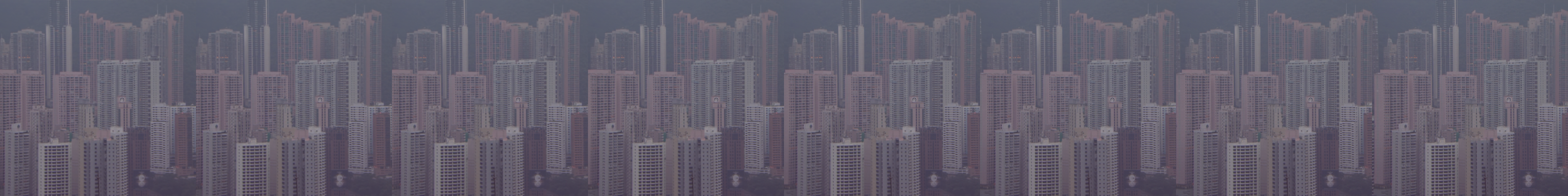}
    \end{minipage}
    
    \vspace{0.3cm}
    
   \begin{minipage}{0.01\textwidth}
        \rotatebox{90}{Ground Truth}
    \end{minipage}%
    \begin{minipage}{0.99\textwidth}
        \centering
        \includegraphics[width=0.96\textwidth]{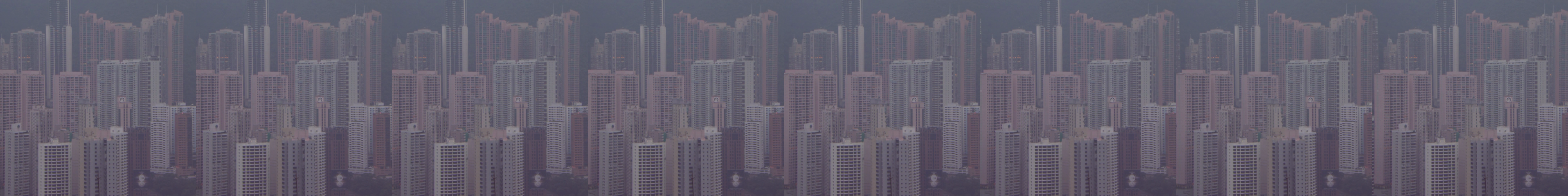}
    \end{minipage}
    \caption{Qualitative comparisons for video \textbf{inpainting}}
    \label{fig:video_comparison_inp}
\end{figure*}
\begin{figure*}
    \begin{minipage}{0.01\textwidth}
        \rotatebox{90}{Measurement}
    \end{minipage}%
    \begin{minipage}{0.99\textwidth}
        \centering
        \includegraphics[width=0.96\textwidth]{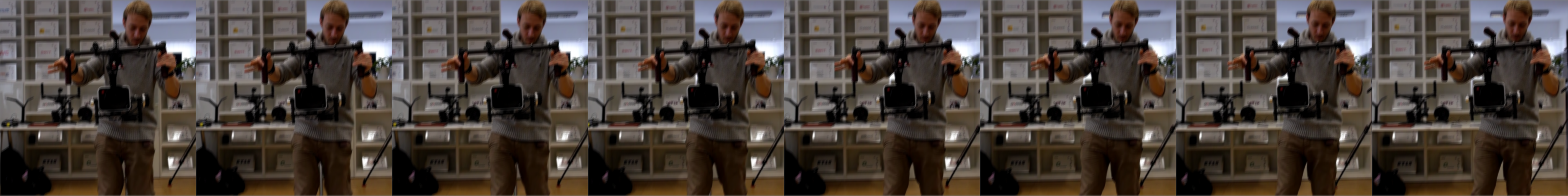}
    \end{minipage}
    
    \vspace{0.3cm}

    \begin{minipage}{0.01\textwidth}
        \rotatebox{90}{DeBlurGANv2}
    \end{minipage}%
    \begin{minipage}{0.99\textwidth}
        \centering
        \includegraphics[width=0.96\textwidth]{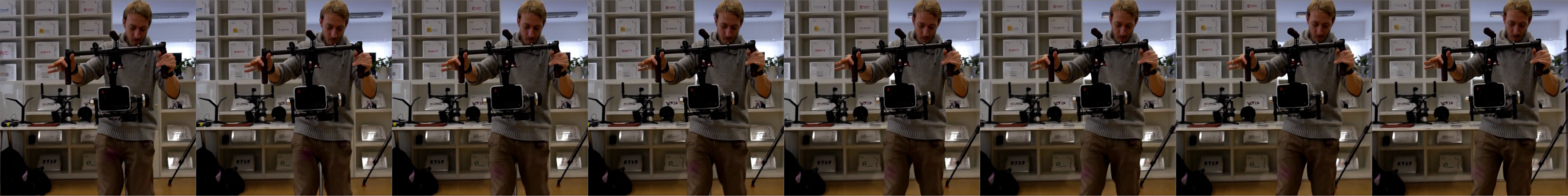}
    \end{minipage}

    \vspace{0.3cm}
    
    \begin{minipage}{0.01\textwidth}
        \rotatebox{90}{Stripformer}
    \end{minipage}%
    \begin{minipage}{0.99\textwidth}
        \centering
        \includegraphics[width=0.96\textwidth]{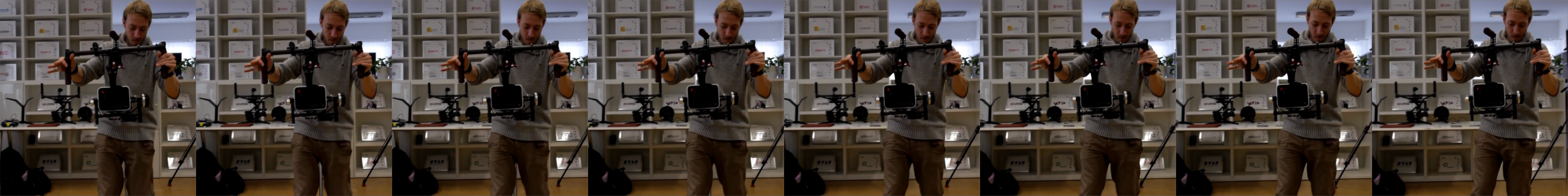}
    \end{minipage}
    
    \vspace{0.3cm}

    \begin{minipage}{0.01\textwidth}
        \rotatebox{90}{SVI}
    \end{minipage}%
    \begin{minipage}{0.99\textwidth}
        \centering
        \includegraphics[width=0.96\textwidth]{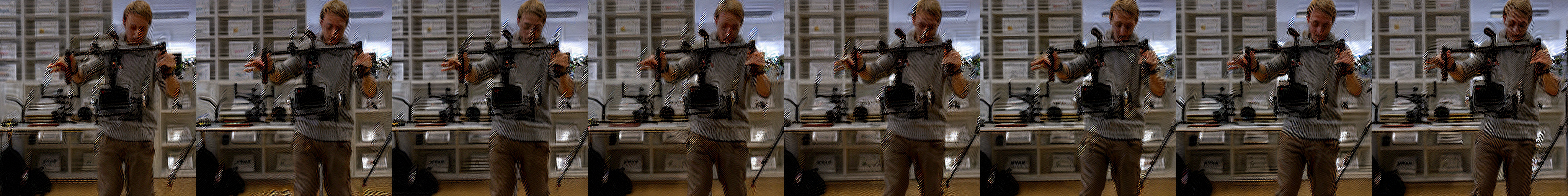}
    \end{minipage}
    
    \vspace{0.3cm}

    \begin{minipage}{0.01\textwidth}
        \rotatebox{90}{VISION-XL}
    \end{minipage}%
    \begin{minipage}{0.99\textwidth}
        \centering
        \includegraphics[width=0.96\textwidth]{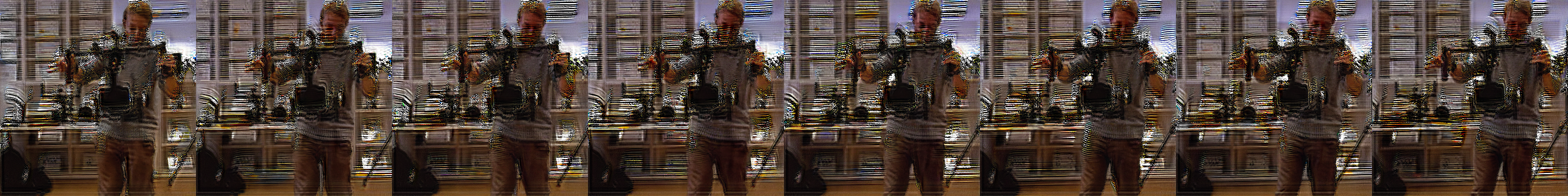}
    \end{minipage}
    
    \vspace{0.3cm}

    \begin{minipage}{0.01\textwidth}
        \rotatebox{90}{Ours}
    \end{minipage}%
    \begin{minipage}{0.99\textwidth}
        \centering
        \includegraphics[width=0.96\textwidth]{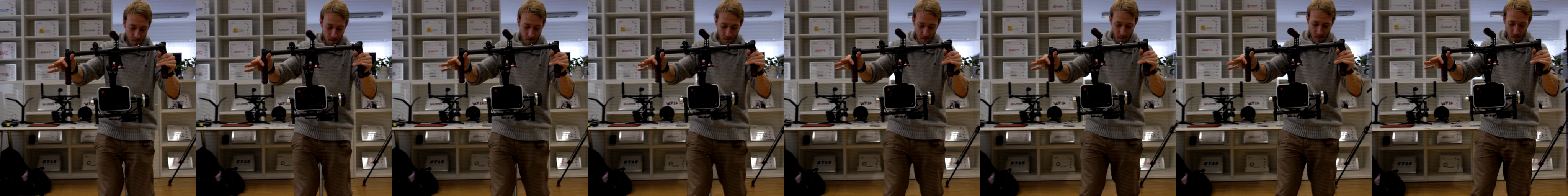}
    \end{minipage}
    
    \vspace{0.3cm}
    
   \begin{minipage}{0.01\textwidth}
        \rotatebox{90}{Ground Truth}
    \end{minipage}%
    \begin{minipage}{0.99\textwidth}
        \centering
        \includegraphics[width=0.96\textwidth]{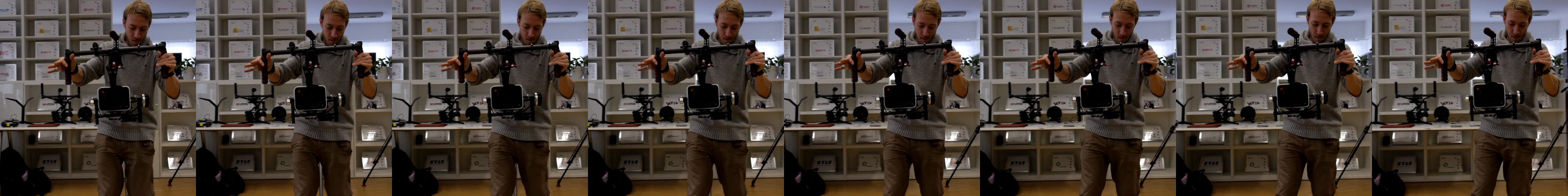}
    \end{minipage}
    \caption{Qualitative comparisons for video \textbf{motion deblurring}}
    \label{fig:video_comparison_motion}
\end{figure*}
\begin{figure*}
    \begin{minipage}{0.01\textwidth}
        \rotatebox{90}{Measurement}
    \end{minipage}%
    \begin{minipage}{0.99\textwidth}
        \centering
        \includegraphics[width=0.96\textwidth]{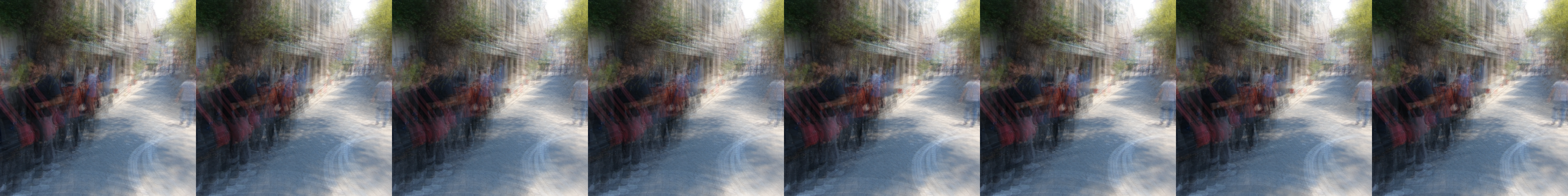}
    \end{minipage}
    
    \vspace{0.3cm}
    
    \begin{minipage}{0.01\textwidth}
        \rotatebox{90}{VRT}
    \end{minipage}%
    \begin{minipage}{0.99\textwidth}
        \centering
        \includegraphics[width=0.96\textwidth]{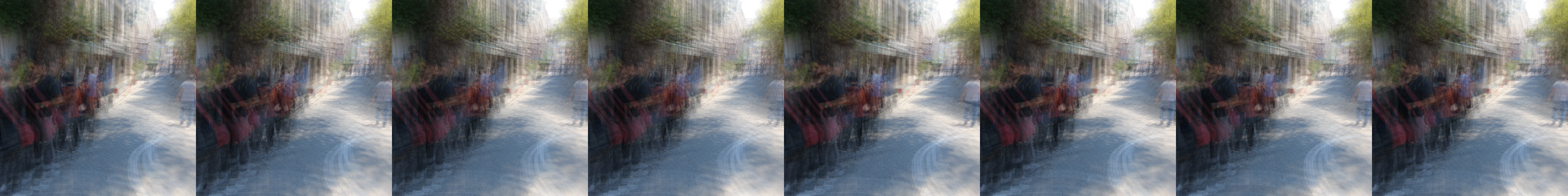}
    \end{minipage}
    
    \vspace{0.3cm}

    \begin{minipage}{0.01\textwidth}
        \rotatebox{90}{SVI}
    \end{minipage}%
    \begin{minipage}{0.99\textwidth}
        \centering
        \includegraphics[width=0.96\textwidth]{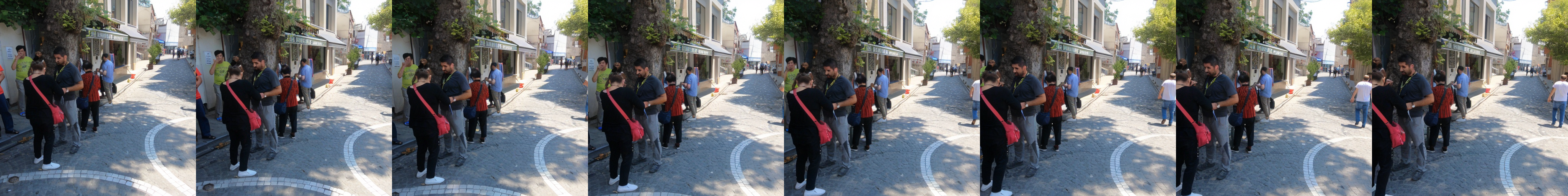}
    \end{minipage}
    
    \vspace{0.3cm}

    \begin{minipage}{0.01\textwidth}
        \rotatebox{90}{VISION-XL}
    \end{minipage}%
    \begin{minipage}{0.99\textwidth}
        \centering
        \includegraphics[width=0.96\textwidth]{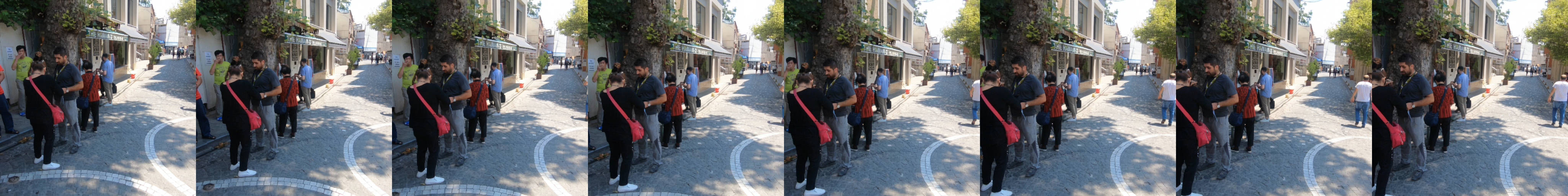}
    \end{minipage}
    
    \vspace{0.3cm}

    \begin{minipage}{0.01\textwidth}
        \rotatebox{90}{Ours}
    \end{minipage}%
    \begin{minipage}{0.99\textwidth}
        \centering
        \includegraphics[width=0.96\textwidth]{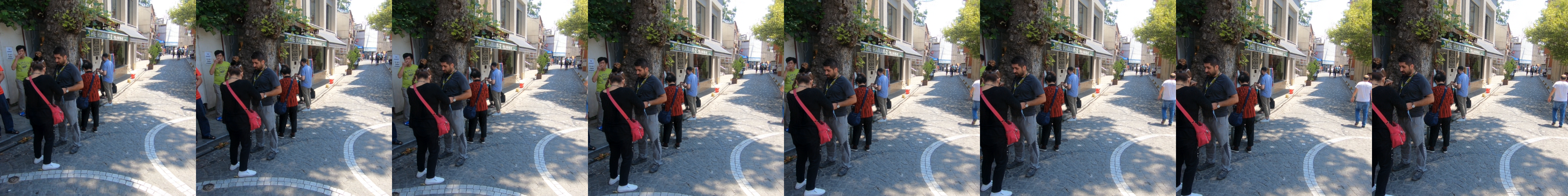}
    \end{minipage}
    
    \vspace{0.3cm}
    
   \begin{minipage}{0.01\textwidth}
        \rotatebox{90}{Ground Truth}
    \end{minipage}%
    \begin{minipage}{0.99\textwidth}
        \centering
        \includegraphics[width=0.96\textwidth]{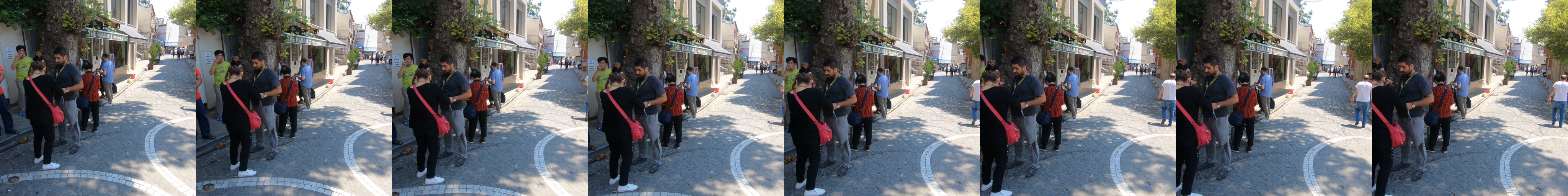}
    \end{minipage}
    \caption{Qualitative comparisons for video \textbf{temporal deconvolution}}
    \label{fig:video_comparison_temporal}
\end{figure*}
\begin{figure*}
    \begin{minipage}{0.01\textwidth}
        \rotatebox{90}{Measurement}
    \end{minipage}%
    \begin{minipage}{0.99\textwidth}
        \centering
        \includegraphics[width=0.96\textwidth]{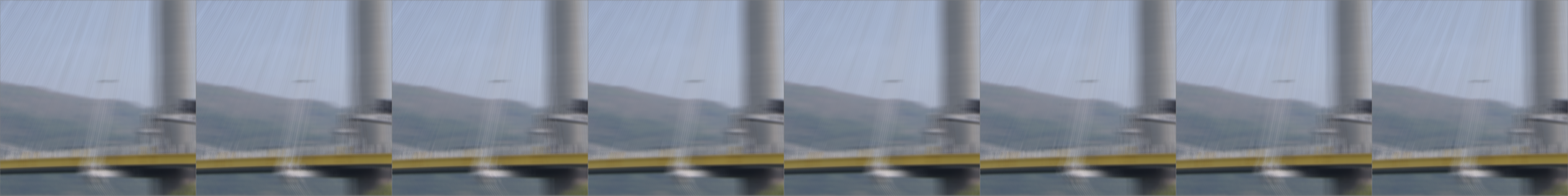}
    \end{minipage}
    
    \vspace{0.3cm}

    \begin{minipage}{0.01\textwidth}
        \rotatebox{90}{DeBlurGANv2}
    \end{minipage}%
    \begin{minipage}{0.99\textwidth}
        \centering
        \includegraphics[width=0.96\textwidth]{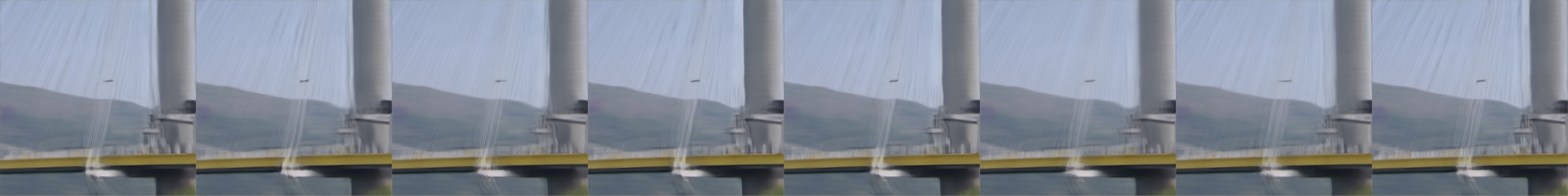}
    \end{minipage}

    \vspace{0.3cm}
    
    \begin{minipage}{0.01\textwidth}
        \rotatebox{90}{Stripformer}
    \end{minipage}%
    \begin{minipage}{0.99\textwidth}
        \centering
        \includegraphics[width=0.96\textwidth]{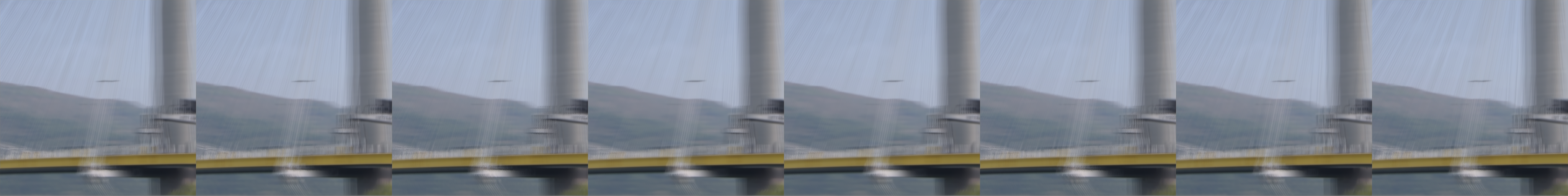}
    \end{minipage}
    
    \vspace{0.3cm}

    

    \begin{minipage}{0.01\textwidth}
        \rotatebox{90}{VISION-XL}
    \end{minipage}%
    \begin{minipage}{0.99\textwidth}
        \centering
        \includegraphics[width=0.96\textwidth]{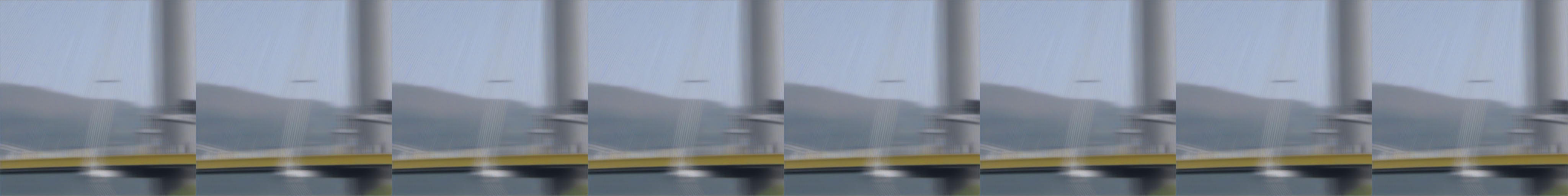}
    \end{minipage}
    
    \vspace{0.3cm}

    \begin{minipage}{0.01\textwidth}
        \rotatebox{90}{Ours}
    \end{minipage}%
    \begin{minipage}{0.99\textwidth}
        \centering
        \includegraphics[width=0.96\textwidth]{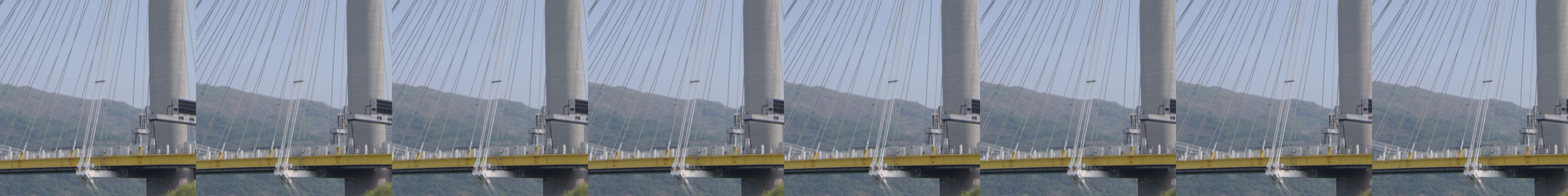}
    \end{minipage}
    
    \vspace{0.3cm}
    
   \begin{minipage}{0.01\textwidth}
        \rotatebox{90}{Ground Truth}
    \end{minipage}%
    \begin{minipage}{0.99\textwidth}
        \centering
        \includegraphics[width=0.96\textwidth]{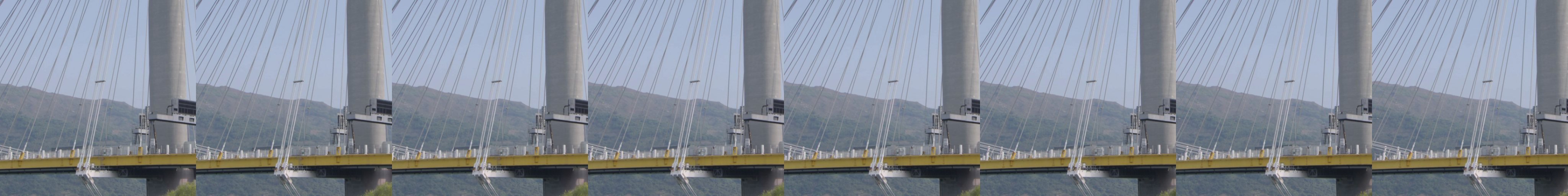}
    \end{minipage}
    \caption{Qualitative comparisons for video \textbf{temporal deconvolution with spatial deblurring}}
    \label{fig:video_comparison_temporal_spatial}
\end{figure*}

We provide more qualitative results for the five video restoration problems, including super-resolution, inpainting, motion deblurring, temporal deconvolution, and temporal deconvolution with spatial deblurring on SPMCS and UDM10 in \cref{fig:comparisons,fig:video_comparison_sr,fig:video_comparison_inp,fig:video_comparison_motion,fig:video_comparison_temporal,fig:video_comparison_temporal_spatial}.

\end{document}